\documentclass[lettersize,journal]{IEEEtran}
\usepackage{amsmath,amsfonts}
\usepackage{algorithmic}
\usepackage{algorithm}
\usepackage{array}
\usepackage[caption=false,font=normalsize,labelfont=sf,textfont=sf]{subfig}
\usepackage{textcomp}
\usepackage{stfloats}
\usepackage{url}
\usepackage{verbatim}
\usepackage{graphicx}
\usepackage{cite}
\usepackage{orcidlink}
\usepackage{multirow}
\usepackage{xspace}
\usepackage{hyperref}
\usepackage{orcidlink}
\usepackage{tabularx}
\definecolor{lightgray}{rgb}{.91,.91,.91}
\definecolor{rouse}{rgb}{0.981,0.961,0.941}
\definecolor{deepred}{rgb}{0.698,0.133,0.133}
\definecolor{blue}{rgb}{0,0,1}
\usepackage{booktabs}
\usepackage{makecell}
\begin{document}

\title{CRISP: Contrastive Residual Injection and Semantic Prompting for Continual Video Instance Segmentation}

\author{Baichen Liu$^{\orcidlink{0000-0002-8844-4794}}$, Qi Lyu$^{\orcidlink{0009-0004-2365-2893}}$, Xudong Wang$^{\orcidlink{0009-0008-3278-0028}}$, Jiahua Dong$^{\orcidlink{0000-0001-8545-4447}}$,\\ 
Lianqing Liu$^{\orcidlink{0000-0002-2271-5870}}$ \IEEEmembership{Senior Member, IEEE}, Zhi Han$^{\orcidlink{0000-0002-8039-6679}}$ \IEEEmembership{Member, IEEE}
\thanks{

This work was supported in part by the National Key Research and Development Program of China under Grant 2024YFB4707700, the National Natural Science Foundation of China under Grant U23A20343, and the Doctoral Research Startup Fund of the Natural Science Foundation of Liaoning Province under Grant 2025-BS-0193. (\textit{Corresponding author: Zhi Han}).

Baichen Liu, Qi Lyu, Xudong Wang, Lianqing Liu, and Zhi Han are with the State Key Laboratory of Robotics and Intelligent Systems, Shenyang Institute of Automation, Chinese Academy of Sciences, Shenyang 110016, China (e-mail: liubaichen@sia.cn; lvqi@sia.cn; wangxudong@sia.cn; lqliu@sia.cn; hanzhi@sia.cn;). Qi Lyu and Xudong Wang are also with the University of Chinese Academy of Sciences, Beijing 100049, China.

Jiahua Dong is with the Mohamed bin Zayed University of Artificial Intelligence, Abu Dhabi, United Arab Emirates. (e-mail: dongjiahua1995@gmail.com).
}
}



\maketitle
\begin{abstract}
Continual video instance segmentation demands both the plasticity to absorb new object categories and the stability to retain previously learned ones, all while preserving temporal consistency across frames. In this work, we introduce Contrastive Residual Injection and Semantic Prompting (CRISP), an earlier attempt tailored to address the instance-wise, category-wise, and task-wise confusion in continual video instance segmentation. For instance-wise learning, we model instance tracking and construct instance correlation loss, which emphasizes the correlation with the prior query space while strengthening the specificity of the current task query. For category-wise learning, we build an adaptive residual semantic prompt (ARSP) learning framework, which constructs a learnable semantic residual prompt pool generated by category text and uses an adjustive query-prompt matching mechanism to build a mapping relationship between the query of the current task and the semantic residual prompt. Meanwhile, a semantic consistency loss based on the contrastive learning is introduced to maintain semantic coherence between object queries and residual prompts during incremental training. For task-wise learning, to ensure the correlation at the inter-task level within the query space, we introduce a concise yet powerful initialization strategy for incremental prompts. Extensive experiments on YouTube-VIS-2019 and YouTube-VIS-2021 datasets demonstrate that CRISP significantly outperforms existing continual segmentation methods in the long-term continual video instance segmentation task, avoiding catastrophic forgetting and effectively improving segmentation and classification performance.
The code is available at \url{https://github.com/01upup10/CRISP}.
\end{abstract}



\section{Introduction}
\IEEEPARstart{D}{eep} learning methods perform very well across a variety of tasks, such as image classification \cite{krizhevsky2012imagenet, guo2021improved,rodriguez2022backdoor,ge2018multi,lin2021structured, 10944285, 10720678}, object detection \cite{wangidentification2024, ren2024learning,wu2021commonality} and image segmentation \cite{11082484,gong2024continual,zhang2022representation,chen2017deeplab,strudel2021segmenter}. Image instance segmentation \cite{dai2016instance,li2017fully,wang2020solo} plays a pivotal role in segmentation tasks by unifying object detection and pixel-level mask prediction, thereby enabling precise delineation of each individual object within an image. In recent years, the methods for image instance segmentation has progressed from multi-stage, cascade-based pipelines to end-to-end and real-time frameworks, and ultimately to unified, attention-driven models that deliver higher accuracy and faster inference. Universal architectures like Mask2Former \cite{cheng2022masked} have extended the attention-driven framework to support panoptic, semantic, and instance image segmentation within a unified model. Furthermore, Mask2Former for video instance segmentation (VIS) also achieves state-of-the-art performance without modifying the architecture \cite{cheng2021mask2formervideo}. Despite these advances, in the process of learning new knowledge, unlike intelligent systems such as humans that can apply previously learned knowledge and skills to new environments, deep networks may encounter the issue of overwriting old class knowledge when learning new information, known as catastrophic forgetting \cite{french1999catastrophic,robins1995catastrophic,thrun1998lifelong}. 

Continual learning (CL) \cite{de2021continual,singh2020calibrating,dong2022federated_FCIL, 10323204} studies how to build a model that can continuously learn new skills from data and tasks, while avoiding catastrophic forgetting of previous tasks. By enabling models to learn incrementally, CL methods \cite{10462930, 10443417, yan2021dynamically} allow a deep neural network to adapt to new tasks while preserving previously learned knowledge. In continual image segmentation tasks, many methods have made significant progress. Knowledge distillation based continual segmentation methods \cite{cermelli2020modeling, douillard2021plop, zhang2022representation} alleviate catastrophic forgetting by transferring knowledge from the old model, while pseudo-labeling allows the new model to train using the previously learned class labels. However, knowledge distillation-based methods incur increased computational overhead due to the need for dual network forwarding and fine-tuning of hyper-parameters, which complicates training. Furthermore, as the number of classes increases, maintaining an efficient and scalable distillation process becomes difficult, and knowledge transfer may struggle to effectively capture new task features \cite{kim2024eclipse}. Prompt tuning based methods \cite{chen2024promptfusion, smith2023coda, wang2023attriclip, wang2022dualprompt, wang2022learning,kim2024eclipse} typically design a shared pool of key-value pairs to manage the prompts learned in each round. The learning criteria for these keys usually involve weak supervision, which pulls selected keys closer to the corresponding query features, with further orthogonality constraints to encourage diversification. The main issue with such methods is that, as the key space is continuously updated and there is no access to past queries or tasks, the selection mechanism itself is subject to catastrophic forgetting, potentially leading to misalignments between queries and keys \cite{menabue2024semantic}.

Continual video instance segmentation (CVIS) introduces unique complexities beyond static image segmentation \cite{cha2021ssul}. Firstly, temporal forgetting \cite{park2021class,villa2022vclimb} arises as the model must retain spatio-temporal reasoning capabilities \cite{pei2023space,cheng2024stsp,chen2025csta} for old classes while learning new ones, and failure to preserve temporal consistency in features (e.g., motion patterns or object trajectories) leads to degraded tracking performance over time. Secondly, incremental updates can distort the latent instance embeddings that are essential for maintaining consistent cross-frame associations. Unlike image tasks where instance identity is frame-local, video models rely on temporally stable embeddings to link instances across frames. Even minor parameter shifts during continual learning can propagate errors in tracking. Thirdly, the absence of annotations for old classes exacerbates the background ambiguity. In dynamic video scenes, unannotated old-class instances (e.g., a previously learned ``car" class in a new ``bus" training phase) risk being misclassified as part of the evolving background, especially when motion cues overlap. Finally, the high dimensionality of the video data (both spatial and temporal \cite{zhao2021video}) challenges conventional continual learning strategies based on rehearsals. Storing representative exemplars from past classes becomes computationally prohibitive, while naive frame sampling often fails to capture essential temporal context for rehearsal. 

\begin{figure*}[!t]
\centering
\includegraphics[width=7in]{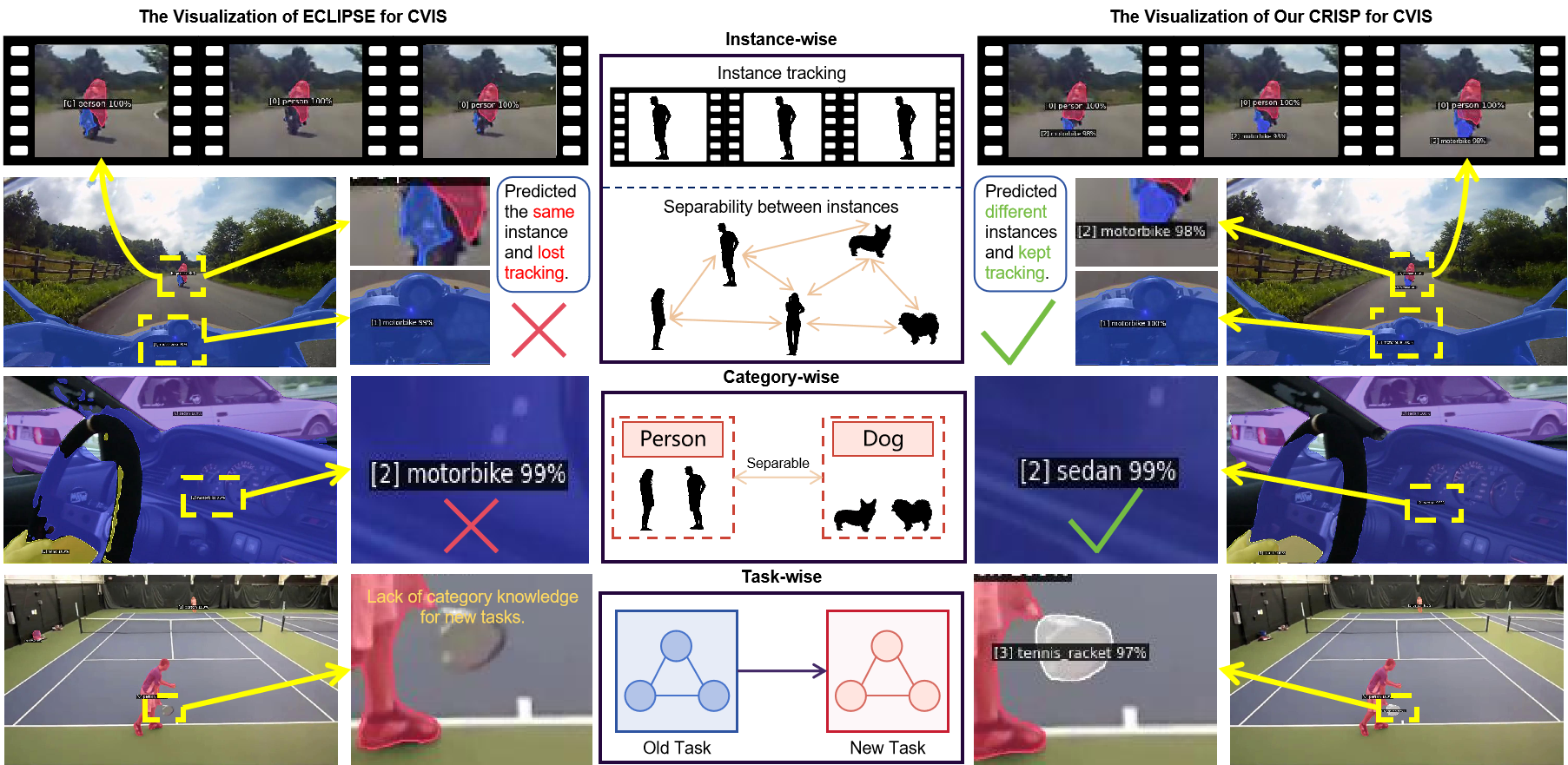}
\caption{Comparison of ECLIPSE \cite{kim2024eclipse} and our CRISP method on CVIS tasks. On the left, ECLIPSE exhibits semantic confusion at instance‑wise, category‑wise, and task‑wise levels, making it difficult to accurately distinguish different instances, leading to misclassification among instances of the same category and frequent misses when learning new classes. On the right, our CRISP framework for CVIS effectively addresses these issues.}
\label{fig:motivation}
\end{figure*}

Since there is currently no method for CVIS, we adapt the state-of-the-art continual image instance segmentation method ECLIPSE \cite{kim2024eclipse} to the video domain and evaluate its performance. ECLIPSE is a distillation‐free framework for continual image segmentation that leverages a frozen backbone and a lightweight prompting mechanism to balance stability and plasticity. By freezing all parameters of the base model and iteratively fine‐tuning only a small set of prompt embeddings as new classes are introduced, ECLIPSE inherently mitigates catastrophic forgetting to some extent while retaining prior knowledge. However, when we apply ECLIPSE to CVIS tasks, as shown in Fig. \ref{fig:motivation}, ECLIPSE exhibits semantic confusion at three levels: instance‑wise, category‑wise, and task‑wise. 
In the first sequence, instance [0] is a person, and instances [1] and [2] are motorbikes. As the video progresses, we observe that ECLIPSE mistakenly merges instances [1] and [2], which are two motorbikes separated by some distance on the road, into a single instance. We term this type of error “instance-wise” confusion.
In the second sequence, instances [0] and [2] are sedans, and instance [1] is a hand. We observe that ECLIPSE sometimes misclassifies instance [2] as a motorbike. We define this type of error, where instances of the same video and category are misclassified into different categories, as “category-wise” confusion.
In the third sequence, the scene contains “person” category from the old task and “tennis\_racket” category from the current task. During continual learning, the number of learnable queries introduced for the new task is fewer than those used initially. To preserve knowledge of old classes, ECLIPSE initializes incremental queries by average pooling and replicating old-class queries. However, this initialization scheme further impairs its ability to learn new categories, causing ECLIPSE to fail to detect the “tennis\_racket” instance [3] of the current task. We refer to this limitation, where the restricted number of learnable variables remains tied to and influenced by the old task, as “task-wise” confusion.

\begin{figure*}[htbp]
\centering
\subfloat[]{\includegraphics[width=1.75in]{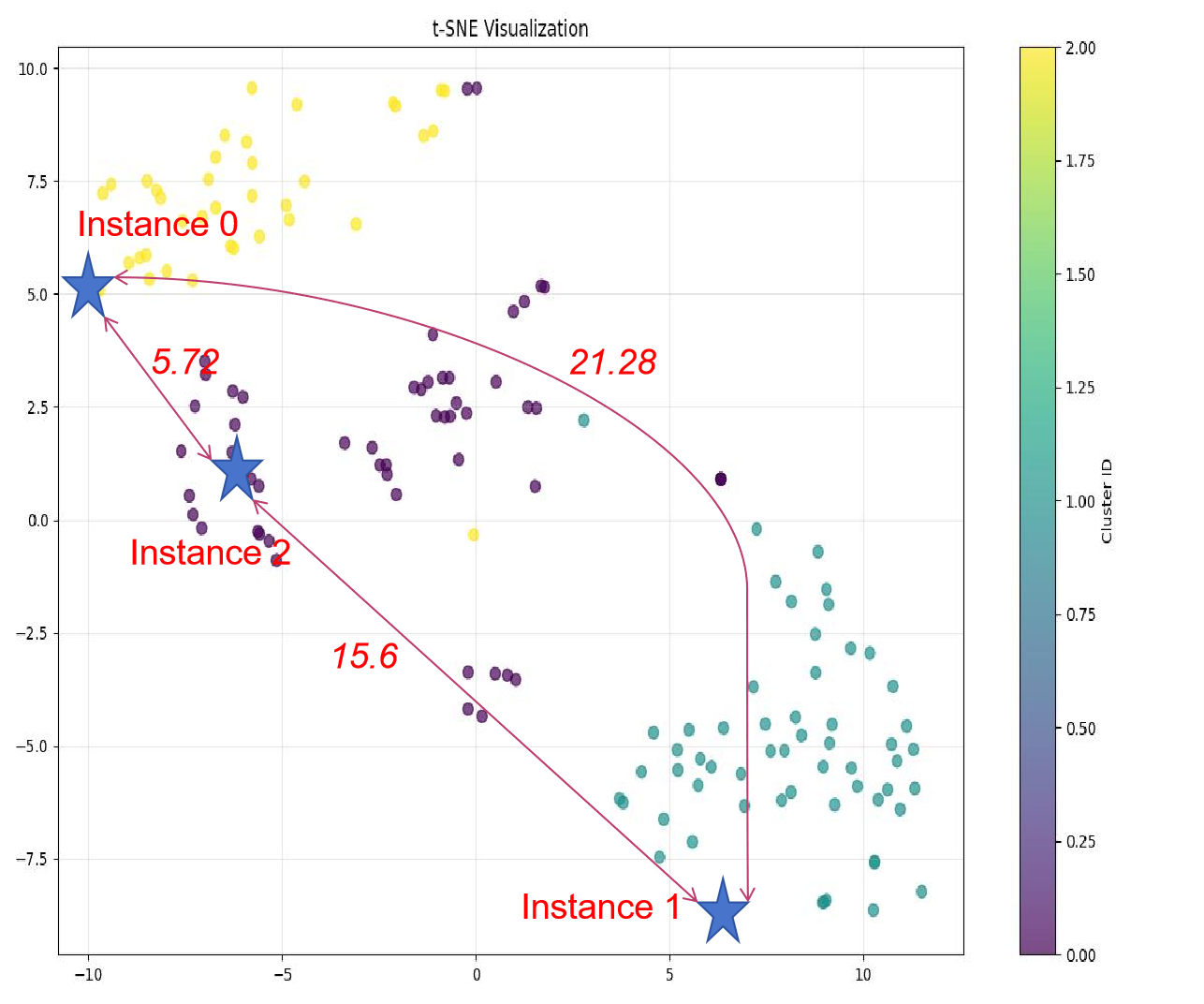}%
\label{fig:ins1}}
\subfloat[]{\includegraphics[width=1.75in]{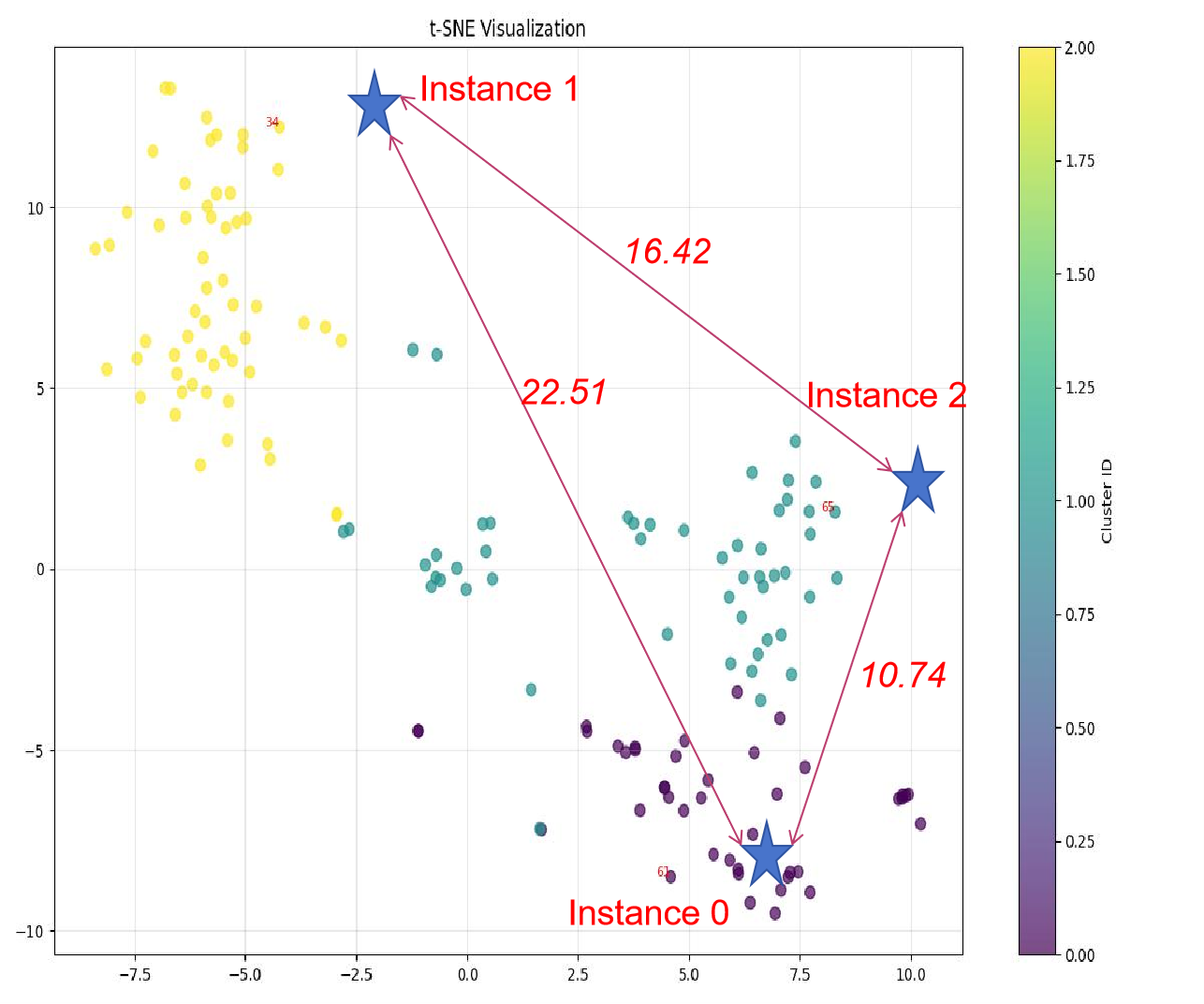}%
\label{fig:ins2}}
\hfil
\subfloat[]{\includegraphics[width=1.75in]{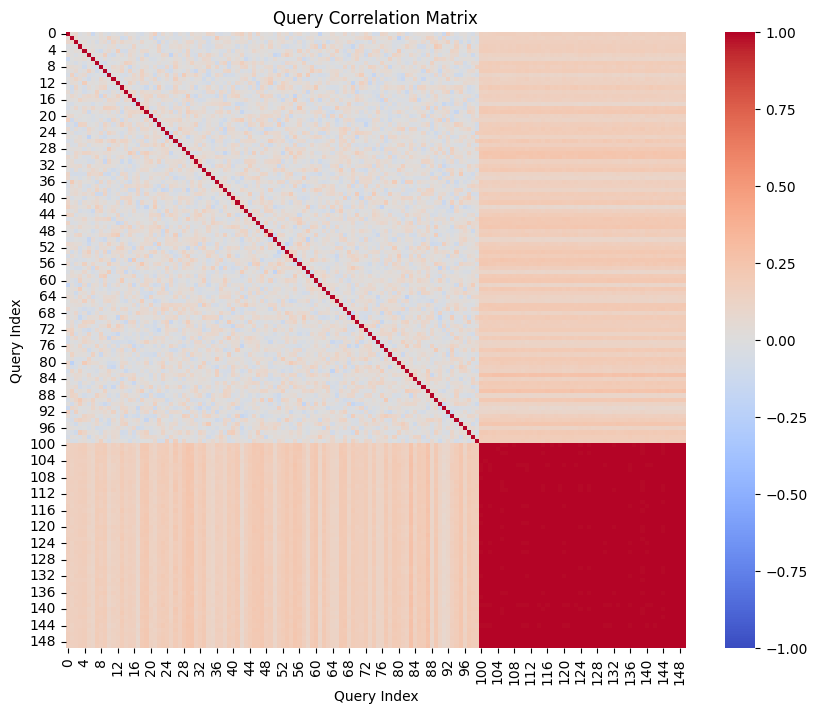}%
\label{fig:query_e}}
\subfloat[]{\includegraphics[width=1.75in]{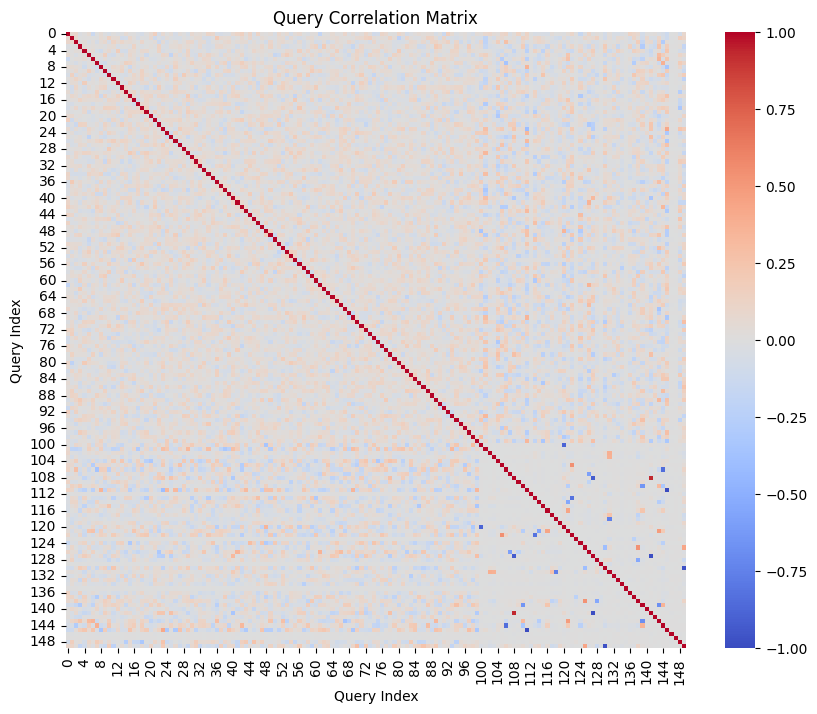}%
\label{fig:query_o}}
\caption{Comparative analysis of query embeddings and their correlations under ECLIPSE \cite{kim2024eclipse} versus our CRISP framework. (a) t‑SNE visualization of the 150 queries for the three instances using ECLIPSE, showing overlapping clusters between sedan and motorbike that explain class confusion. (b) t‑SNE visualization of the same queries under CRISP, illustrating tight, well‑separated clusters for each instance. (c) Covariance matrix of incremental and old‑class queries in ECLIPSE, revealing high inter‑query correlations prone to semantic drift. (d) Covariance matrix for CRISP, demonstrating effective decorrelation and stable embedding relationships that prevent drift.}
\label{fig. ins_query}
\end{figure*}

To further analyze the reasons behind the semantic confusion, we extracted the 150 queries for the three instances from the leftmost image in the second sequence of Fig. \ref{fig:motivation}, and projected them using t‑distributed Stochastic Neighbor Embedding \cite{maaten2008visualizing} (t‑SNE). In addition, we report the pairwise Euclidean distances between the instances. As illustrated in Fig. \ref{fig:ins1}, the sedan and motorbike embeddings lie in close proximity, explaining why ECLIPSE confuses these classes. Additionally, during the continual learning process, we analyze the covariance matrix between the incremental queries and old‐class queries in Fig. \ref{fig:query_e} and find that the incremental queries are highly similar to one another. We hypothesize that this issue stems from the initialization scheme of ECLIPSE for the incremental queries, which generates new category queries by average pooling and replication of old category queries, thereby inducing semantic drift.

Based on the above investigation into semantic confusion at three levels, in this paper, we propose a method called Contrastive Residual Injection and Semantic Prompting (CRISP) to address the challenges in CVIS tasks.
Firstly, for \textbf{instance-wise learning}, considering the specificity between instances, we model instance tracking and construct instance correlation loss, which emphasizes the correlation with the prior query space while strengthening the specificity of the current task query. 
Secondly, for \textbf{category-wise learning}, we build an adaptive residual semantic prompt (ARSP) learning framework, which constructs a learnable residual semantic prompt pool generated by category text and uses an adjustive query-prompt matcher to build a mapping relationship between the query of the current task and the residual semantic prompt. The ARSP module pulls embeddings of the same instance closer and pushes apart those of different classes. The right‑hand side of the second row in Fig. \ref{fig:motivation} shows that our method correctly labels both sedan instances, and Fig. \ref{fig:ins2} demonstrates that the sedan embeddings (instances [0] and [2]) form a tight cluster that is well separated from “hand” embeddings, confirming that our approach substantially improves intra‑class compactness and inter‑class separability. Meanwhile, a semantic consistency loss based on the contrastive learning is introduced to maintain semantic coherence between object queries and residual prompts during incremental training. Thirdly, for \textbf{task-wise learning}, to ensure the correlation at the inter-task level within the query space, we introduce a concise yet powerful initialization strategy for incremental prompts. We treat the prompts from the incremental learning process as query vectors and apply Principal Component Analysis \cite{abdi2010principal,jolliffe2011principal} (PCA) to the collection of old-task queries in order to identify their most significant directions. As shown in Fig. \ref{fig:query_o}, this PCA‑based scheme successfully decorrelates the new incremental queries, and in the right‑hand side of the third row in Fig. \ref{fig:motivation}, our CRISP method correctly labels “tennis\_racker” catogery of the new task. 

In summary, to address instance-wise, category-wise, and task-wise confusion in CVIS, our CRISP method is an earlier attempt applied to CVIS tasks, effectively mitigating the issues of catastrophic forgetting and semantic confusion. The results of the experiment on YouTube-VIS-2019 \cite{Yang2019vis} and YouTube-VIS-2021 \cite{vis2021} show that our proposed CRISP method achieves state-of-the-art performance in CVIS. Furthermore, as the number of continual learning steps increases, CRISP shows significant improvements compared to other continual learning methods. 

The contributions of this paper are as follows:
\begin{itemize}
\item Our proposed CRISP is an earlier attempt applied to CVIS tasks, achieving state-of-the-art results, especially as the number of continual learning steps increases.
\item For instance-wise confusion, we propose an instance correlation loss, which explicitly models instance tracking to enhance the specificity between instances.
\item For category-wise confusion, we propose a one-level residual semantic injection module ARSP, which effectively improves the segmentation performance of the model by injecting residual semantic prompts into the self-attention layer. Besides, we propose a semantic consistency loss based on the contrastive learning to enhance the consistency of similar samples.
\item For task-wise confusion, we propose PCA-guided initialization, which effectively preserves the correlation in the prior query space while ensuring the specificity of the query space of the current task.
\end{itemize}

\section{Related Works}
\textbf{Instance segmentation.} Early image instance segmentation methods such as MNC \cite{dai2016instance} demonstrated the feasibility of combining detection and mask prediction in a cascaded pipeline. FCIS \cite{li2017fully} then advanced this line of work by enabling end‐to‐end training of instance masks alongside classification. The introduction of Mask R‑CNN \cite{he2017mask} unified object detection and mask generation within a two‑stage framework, setting a new standard for accuracy and flexibility. To address real‑time requirements, SOLO \cite{wang2020solo} proposed a location‑based single stage approach that delivers both speed and precision. More recently, transformer‑based architectures like DETR \cite{carion2020end} have reframed image instance segmentation as a direct set prediction problem, and universal frameworks such as Mask2Former \cite{cheng2022masked} have further extended this paradigm by integrating attention mechanisms for panoptic, instance, and semantic segmentation tasks.

VIS is a task that aims to simultaneously detect, segment, and track instance objects in video sequences. ISTR \cite{wang2021end} is the first end-to-end Transformer-based VIS framework, treating VIS task as a direct end-to-end parallel sequence decoding problem. IFC \cite{hwang2021video} consists of a CNN backbone and transformer encoder-decoder layers, and reduces the overhead for information passing between frames by efficiently encoding the context within the input clip, utilizing concise memory tokens as a mean of conveying information as well as summarizing each frame scene. MTN \cite{gu2021class} employs two teacher networks for old and new knowledge to instruct the student network. Specifically, the former teacher network supervises the current student network to preserve the previous knowledge, and the current teacher network supervises the current student network to adapt to new classes. However, the challenge of CVIS task in addressing catastrophic forgetting and background shift remains a cutting-edge research challenge. 

\textbf{Continual image segmentation.} Most continual image segmentation methods typically employ distillation strategies, such as knowledge distillation \cite{cermelli2020modeling, douillard2021plop, zhang2022representation,zhao2022rbc} and pseudo-labeling \cite{cha2021ssul, phan2022class}. Knowledge distillation alleviates catastrophic forgetting by transferring knowledge from the old model, while pseudo-labeling allows the new model to be trained using labels from previously learned categories. These methods have achieved good results in continual image segmentation tasks. However, existing continual segmentation methods, designed for static images, overlook compounding errors from misaligned instance associations across frames, which amplify forgetting in video tasks. Furthermore, temporal feature drift, gradual degradation of motion and appearance representations during incremental updates, is not addressed in image-centric frameworks. Critically, motion cues that could help distinguish background from unannotated old class instances remain underutilized in current continual learning paradigms. 

\textbf{Prompt-based and query-based methods in continual learning.} Prompt-based methods have strong capabilities in preventing catastrophic forgetting in continual learning by learning a small set of insertable model prompts rather than directly modifying the encoder parameters. Query-based methods, on the other hand, leverage a fixed set of learnable object queries to probe the visual features produced by a frozen or partially fine‑tuned encoder. Each query is responsible for attending to and extracting a specific instance-level representation, such as an object mask or bounding box, through cross‐attention mechanisms in Transformer‑based architectures (e.g., DETR \cite{carion2020end}, deformable DETR \cite{zhu2020deformable}, Mask2Former \cite{cheng2022masked}). By decoupling instance extraction from encoder updates, query‑based designs maintain stable feature associations across incremental tasks, helping to preserve temporal and spatial coherence of object instances. In continual segmentation tasks, query‑based and prompt‑based methods are therefore complementary: queries drive instance‑level reasoning over visual inputs of the model, while prompts inject task‑specific or class‑specific semantics that guide the learning process and mitigate forgetting across incremental steps.

The first attempt to use prompt tuning in continual learning is Learning-to-Prompt (L2P) \cite{wang2022learning}, which uses a shared prompt pool across all tasks and selects the most appropriate prompt from the pool using the input image as a query. DualPrompt \cite{wang2022dualprompt} introduces a hierarchy of general and task-specific prompts, employing prefix-tuning instead of traditional prompt tuning, where prefix prompts are added to the keys and values of the MSA layer rather than directly to the input tokens. CODA-Prompt \cite{smith2023coda} further introduces an end-to-end prompt mechanism and orthogonal soft constraints to encourage the independence of the prompts. STAR-Prompt \cite{menabue2024semantic} proposes a two-level adaptive mechanism that utilizes stable class prototypes from the CLIP model and query images as keys to retrieve prompts, enhancing the stability of the prompt selection strategy. CoMFormer \cite{cermelli2023comformer} introduced the first query‑centric approach to continual panoptic segmentation, combining knowledge distillation with pseudo‑labeling to alleviate catastrophic forgetting. Building on this foundation, CoMasTRe \cite{gong2024continual} retains the distillation objective but disentangles mask generation from class prediction within the continual learning pipeline. In a similar vein, BalConpas \cite{chen2024strike} combats forgetting by uniting feature‑level distillation with a replay buffer of representative samples, allowing the model to acquire new classes without compromising already learned knowledge. 

\section{METHODOLOGY}
\begin{figure*}[!t]
\centering
    \includegraphics[width=7.15in]{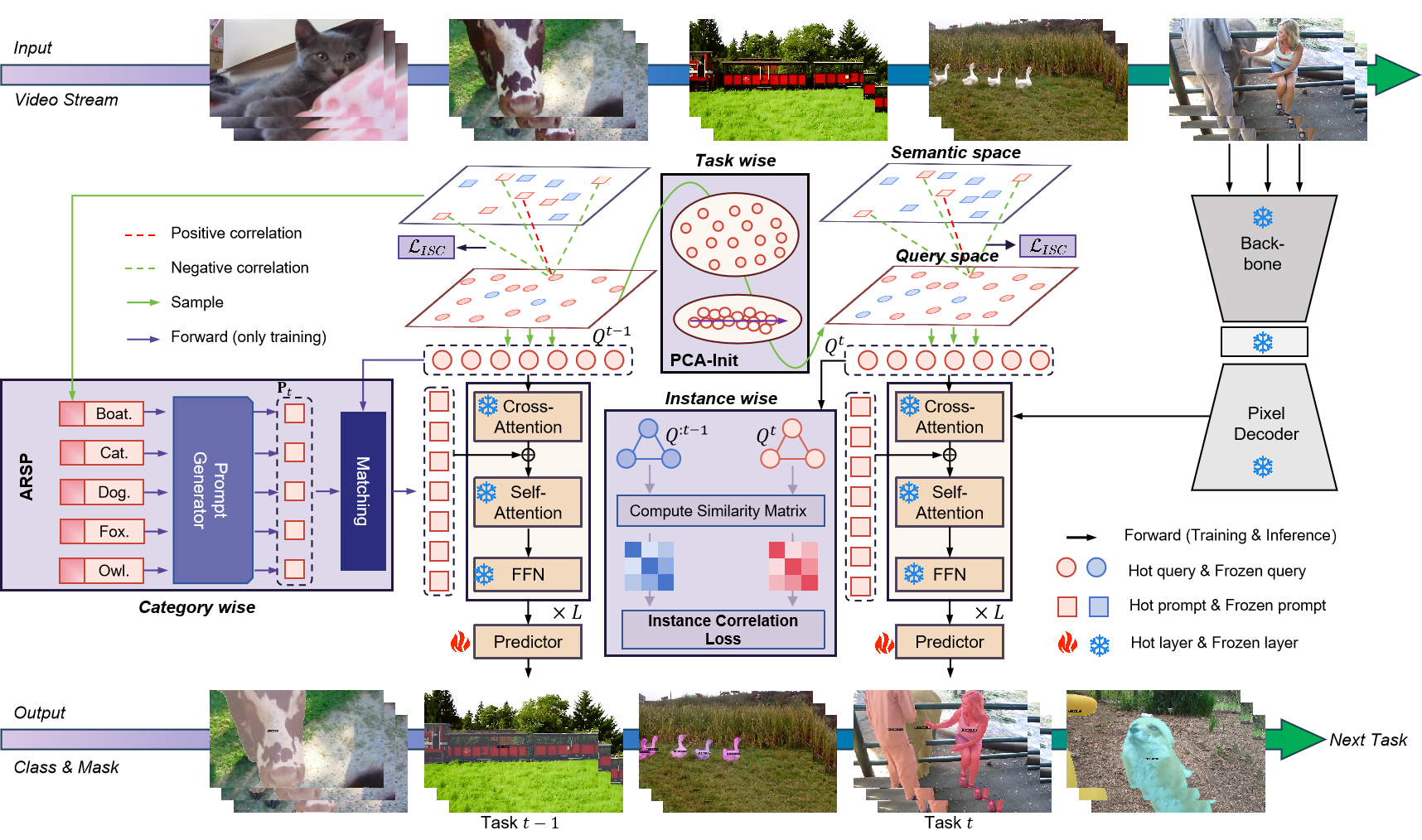}
    \caption{Our proposed CRISP architecture for CVIS. An ARSP module (sec \ref{arsp}) with instance semantic consistency loss (sec \ref{isc}) for category-wise, a PCA-guided initiazation (sec \ref{pi}) for task-wise, and an instance correlation loss (\ref{ic}) for instance-wise. It is worth noting that our method only injects residual semantic prompts during training, and the queries and prompts of old tasks would be frozen.}
    \label{fig:method}
\end{figure*}
\subsection{Problem Setting}

The goal of CVIS is to learn a model capable of segmenting and tracking object instances across video sequences while incrementally adapting to new classes introduced over multiple learning phases. Let $\mathcal{V}$ denote the input video space, where each video $v \in \mathcal{V}$ consists of a temporal sequence of frames $\{x_1, \ldots, x_T\}$. At each learning step $t$, a new set of classes $\mathcal{C}_t$ is introduced, and the model needs to segment and associate instances between frames for all observed classes $\mathcal{C}_{1:t}$. During step $t$, the model receives a training dataset $\mathcal{D}_t$ that contains videos with annotations only for $\mathcal{C}_t$. Each annotation comprises instance masks $m_{i,k}^t \in \{0,1\}^{H \times W}$ and class labels $c_{i,k}^t \in \mathcal{C}_t$ for the $k^{th}$ object in frame $x_i$, where $H \times W$ is the spatial resolution. Critically, annotations for previous classes $\mathcal{C}_{1:t-1}$ and future classes $\mathcal{C}_{t+1:T}$ are \textit{entirely absent}, not even collapsed into a background class. The model must predict per-frame instance masks and class labels for $\mathcal{C}_{1:t}$, maintain temporal consistency by associating identical instances across frames, and preserve performance on $\mathcal{C}_{1:t-1}$ while learning $\mathcal{C}_t$ to avoid catastrophic forgetting.

\subsection{Overview}

Firstly, category-wise confusion is inevitable due to the inherent visual similarities across tasks. To mitigate this issue, we propose an adaptive residual semantic prompt (ARSP) learning framework, complemented by an instance semantic consistency loss to enhance discriminative feature learning.
Secondly, while prompt tuning partially alleviates catastrophic forgetting by freezing most parameters, it still suffers from task-wise confusion. To address this limitation, we introduce PCA-guided initialization, a novel approach that allows new task queries to effectively inherit prior knowledge from previous tasks, thereby improving task discrimination.
Thirdly, to tackle instance-wise confusion, we propose an instance association loss, grounded in our instance tracking modeling, which strengthens the consistency of instance representations across tasks.

\subsection{Adaptive Residual Semantic Prompts for Training}
\label{arsp}
ARSP is a one-level Prompt-tuning strategy to guide the model to learn semantic signals during training to address category-wise confusion. As shown in Fig. \ref{fig:method}, ARSP integrates three core components: a residual semantic prompt generator, an adaptive query-prompt matching mechanism, and a hierarchical injection architecture.

\textbf{Residual Semantic Prompt Generator.} Following CoOp \cite{zhou2022learning} paradigm, we construct a residual semantic prompt pool $\mathbf{P}_t=[\mathbf{p}_0|\mathbf{p}_1|\mathbf{p}_2|...|\mathbf{p}_{c_t}] \in \mathbb{R}^{c_t \times d}$, which contains $c_t$ class-specific prototypes in a $d$ dimensional space, where $c_t$ corresponds to the total number of categories in task $t$. A prompt generator $\mathcal{G}_ \theta$, implemented on the CLIP \cite{radford2021learning} text encoder, maps the learnable token sequences $\mathcal{X} \in \mathbb{R}^{c_t \times d}$ to the residual semantic prompts $\mathbf{P}_t$ through $\mathcal{G}_ \theta$.

\textbf{Adaptive Query-prompt Matching.} We obtain the residual semantic prompts corresponding to each query ${q}^{l} \in \mathbb{R}^{d}$ through a matching mechanism. The matching process can be denoted as:
\begin{equation} 
\label{match_processing} 
    \begin{aligned}
    &\mathbf{a} \!=\! 
    ( a_1, a_2, \cdots, a_{c_t} )^T,&\\ 
    \text{s.t.} \!\verb| | a_i \!= \! \arg &\max_{j=1}^{c_t} S_{i,j} \verb| |
    \text{for } \! \verb| | i = 1, 2, \dots, N^t_q, & \\
   & S_{i,j} \! = \! \frac{\mathbf{Q}_t\mathbf{P}^T_t}{\parallel\!\mathbf{Q}_t\! \parallel_F \mathbf \parallel \mathbf{P}^T_t \!\parallel_F},& 
    \end{aligned} 
\end{equation}
where $N^t_q$ indicates the number of queries in task $t$, $\mathbf{Q}_t \in \mathbb{R}^{N^t_q \times d}$ represents object queries for task \textit{t} with dimension \textit{d}, and $\parallel  \!\cdot\! \parallel$ denotes frobenius norm.

\textbf{Hierarchical Injection Architecture.} Building upon Mask2Former for video \cite{cheng2021mask2formervideo} architecture, we inject residual prompts into multi-scale feature decoding through a modified self-attention operation:
\begin{equation}
    \mathbf{O}^l = Softmax(\frac{\mathbf{Q}^{l}(\mathbf{K}^{l})^T} {\sqrt{ d_{k} }})(\mathbf{V}^{l} + \mathbf{P}^{l}_{m}),
\end{equation}
where $\mathbf{O}^l$ represents the output of self-attention $l$-th layer, $\mathbf{Q}^l$, $\mathbf{K}^l$ and $\mathbf{V}^{l}$ denote the query, key and value of the Transformer decoder layer $l$, respectively, and $\mathbf{P}^{l}_{m}$ denotes the matched residual semantic prompts corresponding to $\mathbf{Q}^l$.

\subsection{Instance Semantic Consistency Loss}
\label{isc}
During the inference process, we do not inject residual semantic prompts for faster inference speed. Therefore, to maintain semantic consistency between object queries and residual prompts during continual training, we formulate a semantic consistency loss based on the contrastive learning framework proposed in \cite{lee2024context}, which aligns the semantic space with the query space. Given object queries $\mathbf{Q}_t \in \mathbb{R}^{N^t_q \times d}$ and the residual prompts of the current task $\mathbf{P}_t \in \mathbb{R}^{c_t \times d}$, by calculating the cosine similarity matrix $\mathbf{S}\in \mathbb{R}^{N^t_q \times c_t}$, the instance semantic consistency loss is denoted as:
\begin{equation}
     \mathcal{L}_{\text{ISC}} = \frac{1}{N^t_q} \sum_{i=1}^{N^t_q} \log\left(1 + \frac{\sum_{j=1}^{c_t} \mathbb{I}_{ij} \cdot \exp(\mathbf{S}_{ij})}{\exp(\mathbf{S}_{i, {a}_i})}\right),
\end{equation}
where $a_i$ represents the index derived from applying Eq. \ref{match_processing} to the object query $q^l \in Q^l$, and $\mathbb{I}_{ij}$ is denoted as:
\begin{equation}
    \mathbb{I}_{ij} =
\begin{cases}
0 & \text{if } j = a_i \\
1 & \text{otherwise.}
\end{cases}
\end{equation}

\subsection{PCA-guided Initialization}
\label{pi}
Although ECLIPSE\cite{kim2024eclipse} achieves excellent performance in image segmentation tasks by generating queries for new categories $\mathcal{C}^t$ through average pooling and replication operations, its direct adaptation to video segmentation tasks exhibits notable limitations. Specifically, due to the high consistency among the generated queries shown in Fig. \ref{fig:query_e}, the characteristic significantly affects the separability between tasks in long-term video instance segmentation scenarios, leading to task-wise confusion. To alleviate task-wise confusion caused by query consistency during training, we propose a concise and effective prompt initialization strategy. Firstly, Principal Component Analysis \cite{jolliffe2011principal} (PCA) is applied to extract the principal components of the queries from the old task. Subsequently, we sample principal component vectors of the top-$c_t$ corresponding largest eigenvalues. Finally, these sampled latent representations are aligned with the prior query manifold to form the initial query for the current task. The initialization process is presented in detail in Alg. \ref{alg: PCA_init}.

\begin{algorithm}[h!]
\caption{PCA-guided initialization.}
\begin{algorithmic}[1]
\REQUIRE
queries of old tasks $\mathbf{Q}_o\in \mathbb{R}^{N^{:t-1}_q \times d}$, the length of queries from previous tasks $N^{:t-1}_q$, total categories in old tasks $c_{:t-1}$, total categories in current task $c_{t}$.
\ENSURE
initialized queries for current task $\mathbf{Q}_t$.
\STATE $ \mathbf{Q}' \gets PCA(\mathbf{Q}_o) \in \mathbb{R}^{c_{:t-1} \times d}$;
\STATE $\mathcal{N} \gets [\parallel q'_1  \parallel_2 ,  \parallel q'_2  \parallel_2 , \parallel q'_3  \parallel_2 , \cdots , \parallel q'_{c_{:t-1}}  \parallel_2 ]$;
\STATE \textbf{Select indices of top-$c_t$ norms:}
\STATE $ \mathcal{I} \gets argsort(-\mathcal{N})[:c_t]$; 
\STATE $ \mathbf{Q}_t \gets \mathbf{Q}'[\mathcal{I},:]\in \mathbb{R}^{c_t \times d}  $;
\STATE \textbf{Align} $\mathbf{Q}_t$:
\STATE $ a_{ori} \gets \frac{1}{N^{:t-1}_q}\sum^{N^{:t-1}_q}_ {i=1}\parallel \mathbf{Q}_o[i,:]\parallel_2 $;
\STATE $a_{pca} \gets \frac{1}{c_{:t-1}}\sum^{c_{:t-1}}_ {i=1}\mathcal{N}$;
\STATE $ \mathbf{Q}_t \gets \frac{a_{ori}}{a_{pca}}\mathbf{Q}_t  $;
\STATE \textbf{return} $\mathbf{Q}_t$.
\end{algorithmic}
\label{alg: PCA_init}
\end{algorithm}

\subsection{Instance Correlation Loss}
\label{ic}
In CVIS tasks, we classify the challenges of cross-frame instance tracking and the confusion between different instances within the same frame as instance-wise confusion.
Mask2Former for video \cite{cheng2021mask2formervideo} introduces object queries to model the correlation between instances by representing different instances with a series of queries. Assuming the model has a total of $N$ queries, it can theoretically track $N$ instances. Therefore, these queries should be pairwise separable. To this end, we model instance tracking as:
\begin{equation}
\label{eq:minmax}
     min\parallel \hat{\mathbf{Q}}\hat{\mathbf{Q}}^T-\mathbf{\textit{I}} \parallel _{F},
\end{equation}
where $\hat{\mathbf{Q}}$ represents the normalized query and $\mathbf{\textit{I}}$ is the identity matrix.

Considering that different instances within the same category should possess certain similarity, queries are not completely orthogonal to each other. Therefore, we characterize the correlation between queries using the inner product matrix of $\mathbf{Q}$ and $\mathbf{Q}^T$ and model it as follows:
\begin{equation}
\label{eq:minmax2}
     min\parallel \hat{\mathbf{Q}}\hat{\mathbf{Q}}^T-{\hat{\mathbf{Q}}_{0}}{\hat{\mathbf{Q}}_{0}^T} \parallel _{F}.
\end{equation}
Based on Eq. \ref{eq:minmax2}, we propose an instance correlation loss function, denoted as:
\begin{equation}
    \mathcal{L}_{IC} = MSE(\hat{\mathbf{Q}}_{t}\hat{\mathbf{Q}}_{t}^T,{\hat{\mathbf{Q}}_{0}}{\hat{\mathbf{Q}}_{0}^T}),
\end{equation}
where $\hat{\mathbf{Q}}_t$ denotes the normalized query for task $t$, $\hat{\mathbf{Q}}_0$ denotes the normalized query for the initial task, and $MSE(\cdot)$ represents the mean squared error loss.

\subsection{Total  Loss}

We follow \cite{cheng2022masked} to define the segmentation loss function $\mathcal{L}_{Seg}$, which consists of a mask loss composed of binary cross-entropy loss and dice loss, as well as a classification loss defined by cross-entropy. The total loss function can be donated as:
\begin{equation}
    \label{eq:total_loss}
    \mathcal{L}_{tol} = \mathcal{L}_{Seg} + \lambda_{ISC}\mathcal{L}_{ISC} +\lambda_{IC}\mathcal{L}_{IC}.
\end{equation}
In our method, we set $\lambda_{ISC}=\lambda_{IC}=3$. We also add auxiliary instance correlation losses $\mathcal{L}^l_{IC}$ at each transformer decoder layer $(l\in\{1,\ldots,L\})$.
\begin{table}[]
\centering
\caption{CVIS results on YouTube-VIS-2019 dataset. The results of the 20-4 scenario are shown above, and the results of the 20-2 scenario are shown below. The optimal result is highlighted in red, and the sub-optimal result is highlighted in blue. The following tables use the same notation.}
\begin{tabular}{ccccccc}\toprule
  \textbf{Model} & \textbf{mAP}& \textbf{AP50}& \textbf{AP75}& \textbf{AR1}& \textbf{AR10}& \textbf{FR}\\\midrule
                                                   FT             & 6.53& 11.39& 6.96& 11.59& 11.99&                                         21.19\\
                                                   MiB\cite{Cer2020MiB}            & 9.01& 17.75& 7.29& 14.03& 24.34&                                         17.78\\
                                                   CoMFormer \cite{cermelli2023comformer}      & 12.76& 24.43& 11.08& 16.85& 26.22& 13.6\\
                                                   ECLIPSE\cite{kim2024eclipse}        & \textbf{\textcolor{blue}{25.03}}& \textbf{\textcolor{blue}{39.83}}& \textbf{\textcolor{blue}{27.59}}& \textbf{\textcolor{blue}{32.81}}& \textbf{\textcolor{blue}{40.48}}&                                         \textbf{\textcolor{blue}{2.23}}\\
                                                   CoMBO\cite{fang2025combo}          & 15.8& 29.40& 16.6& 25.54& 28.55&                                         16.05\\
                          CRISP           & \textbf{\textcolor{deepred}{28.1}}& \textbf{\textcolor{deepred}{43.3}}& \textbf{\textcolor{deepred}{31.98}}& \textbf{\textcolor{deepred}{38.04}}& \textbf{\textcolor{deepred}{44.71}}&                                         \textbf{\textcolor{deepred}{1.93}}\\ \midrule
                                                   FT             & 3.72& 6.61& 4.27& 9.67& 10.13&                                         15.56\\
                                                   MiB\cite{Cer2020MiB}            & 0.94& 2.07& 0.91& 3.58& 5.86&                                         16.26\\
                                                   CoMFormer \cite{cermelli2023comformer}      & 5.26& 9.01& 5.32& 9.57& 12.01&                                         11.4\\
                                                   ECLIPSE\cite{kim2024eclipse}        & \textbf{\textcolor{blue}{19.09}}& \textbf{\textcolor{blue}{29.05}}& \textbf{\textcolor{blue}{21.81}}& \textbf{\textcolor{blue}{26.06}}& \textbf{\textcolor{blue}{31.92}}&                                         \textbf{\textcolor{blue}{3.21}}\\
                                                   CoMBO\cite{fang2025combo}          & 4.37& 8.17& 4.25& 9.59& 10.71&                                         15.02\\
  CRISP           & \textbf{\textcolor{deepred}{25.14}}& \textbf{\textcolor{deepred}{37.46}}& \textbf{\textcolor{deepred}{28.17}}& \textbf{\textcolor{deepred}{32.87}}& \textbf{\textcolor{deepred}{39.13}}&                                         \textbf{\textcolor{deepred}{1.13}}\\ \bottomrule
\end{tabular}
\label{tab:exp_table0}
\end{table}

\begin{table}[t!]
\centering
\caption{CVIS results on YouTube-VIS-2021 dataset. The results of the 20-5 scenario are shown above, and the results of the 10-10 scenario are shown below.}
\begin{tabular}{ccccccc}\toprule
   \textbf{Model} & \textbf{mAP}& \textbf{AP50}& \textbf{AP75}& \textbf{AR1}& \textbf{AR10}&\textbf{FR}\\ \midrule
                                                   FT             & 4.08& 6.90& 4.52& 6.43& 7.36&                                         32.63\\
                                                   MiB\cite{Cer2020MiB}            & 5.82& 10.87& 4.49& 9.52& 15.96&                                         25.44\\
                                                   CoMFormer \cite{cermelli2023comformer}      & 14.49& 28.01& 14.46& 23.01& 30.15&                                         10.58\\
                                                   ECLIPSE\cite{kim2024eclipse}        & \textbf{\textcolor{blue}{22.26}}& \textbf{\textcolor{blue}{33.39}}& \textbf{\textcolor{blue}{24.74}}& \textbf{\textcolor{blue}{24.81}}& \textbf{\textcolor{blue}{31.03}}&                                         \textbf{\textcolor{blue}{5.03}}\\
                                                   CoMBO\cite{fang2025combo}          &                         15.35&                         25.18&                         16.40&                         24.01&                         29.39&                                         19.88\\
                          CRISP           & \textbf{\textcolor{deepred}{26.13}}& \textbf{\textcolor{deepred}{39.13}}& \textbf{\textcolor{deepred}{28.82}}& \textbf{\textcolor{deepred}{28.84}}& \textbf{\textcolor{deepred}{37.49}}&                                         \textbf{\textcolor{deepred}{4.38}}\\ \midrule 
FT             & 5.12& 10.24& 4.54& 7.97& 9.34&                                         42.96\\
                                                   MiB\cite{Cer2020MiB}            & 8.14& 13.03& 9.06& 10.11& 12.25&                                         24.95\\
                                                   CoMFormer \cite{cermelli2023comformer}      & 12.83& 23.35& 13.19& 16.92& 26.14&                                         23.67\\
                                                   ECLIPSE\cite{kim2024eclipse}        & 19.04& 30.94& 19.92& \textbf{\textcolor{blue}{25.4}}& 32.06&                                         \textbf{\textcolor{blue}{4.38}}\\
                                                   CoMBO\cite{fang2025combo}          &                         \textbf{\textcolor{deepred}{20.29}}&                         \textbf{\textcolor{deepred}{33.94}}&                         \textbf{\textcolor{deepred}{21.58}}&                         \textbf{\textcolor{deepred}{26.19}}&                         \textbf{\textcolor{deepred}{33.64}}&                                         19.75\\
  CRISP           &                         \textbf{\textcolor{blue}{20.05}}&                         \textbf{\textcolor{blue}{31.82}}&                         \textbf{\textcolor{blue}{21.54}}&                         25.26&                         \textbf{\textcolor{blue}{32.71}}&                                        
\textbf{\textcolor{deepred}{2.73}}\\ \bottomrule \end{tabular}

\label{tab:exp_table}
\end{table}

\begin{table*}[]
\centering
\caption{20-4 results on YouTube-VIS-2019 dataset. Each cell in the table represents APs, APm, and APl, respectively. This setting is applied consistently across Tab. \ref{tab:task-wise-20-4}-\ref{tab:task-wise-10-10}. }
\begin{tabular}{>{\centering\arraybackslash}p{0.12\linewidth}>{\centering\arraybackslash}p{0.12\linewidth}>{\centering\arraybackslash}p{0.12\linewidth}>{\centering\arraybackslash}p{0.12\linewidth}>{\centering\arraybackslash}p{0.12\linewidth}>{\centering\arraybackslash}p{0.12\linewidth}>{\centering\arraybackslash}p{0.12\linewidth}}\toprule
   \textbf{Model} & \textbf{Step0}& \textbf{Step1}& \textbf{Step2}& \textbf{Step3}& \textbf{Step4}&\textbf{Step5}\\ \midrule
                                                   FT             & 12.20/17.64/21.07& 9.34/15.86/19.49& 6.01/10.39/12.23& 6.31/9391/15.17& 3.96/5.03/10.25&                                         4.03/4.25/10.81\\
                                                   MiB\cite{Cer2020MiB}            & 10.82/16.63/22.15& 9.9/13.54/20.43& 6.58/17.52/21.12& 6.48/10.74/21.13& 5.12/12.52/19.71&                                         3.4/10.83/19.02\\
                                                   CoMFormer \cite{cermelli2023comformer}      & 10.34/16.42/19.57& 9.26/17.65/22.09& 9.76/17.61/21.83& 8.84/16.26/24.36& 7.07/13.23/24.87&                                         8.32/13.5/20.24\\
                                                   ECLIPSE\cite{kim2024eclipse}        & \textbf{\textcolor{blue}{11.96}}/\textbf{\textcolor{blue}{22.03}}/27.86& 9.75/19.92/26.63& 10.28/\textbf{\textcolor{blue}{20.89}}/28.18& \textbf{\textcolor{blue}{10.88}}/\textbf{\textcolor{blue}{22.88}}/32.19& \textbf{\textcolor{blue}{10.88}}/\textbf{\textcolor{blue}{22.77}}/\textbf{\textcolor{blue}{32.99}}&         \textbf{\textcolor{blue}{12.76}}/\textbf{\textcolor{blue}{24.96}}/\textbf{\textcolor{blue}{35.65}}\\
                                                   CoMBO\cite{fang2025combo}          &                         \textbf{\textcolor{deepred}{13.16}}/\textbf{\textcolor{deepred}{22.51}}/\textbf{\textcolor{blue}{28.72}}&                         \textbf{\textcolor{deepred}{12.34}}/\textbf{\textcolor{deepred}{22.20}}/\textbf{\textcolor{deepred}{29.87}}&                         \textbf{\textcolor{blue}{10.67}}/19.78/\textbf{\textcolor{deepred}{30.49}}&                         10.56/18.97/\textbf{\textcolor{deepred}{33.66}}&                         8.62/18.94/29.10&                                         6.44/14.5/26.16\\
                          CRISP           & \textbf{\textcolor{blue}{11.96}}/21.96/\textbf{\textcolor{deepred}{29.03}}& \textbf{\textcolor{blue}{11.38}}/\textbf{\textcolor{blue}{21.78}}/\textbf{\textcolor{blue}{28.16}}& \textbf{\textcolor{deepred}{12.32}}/\textbf{\textcolor{deepred}{22.89}}/\textbf{\textcolor{blue}{29.61}}& \textbf{\textcolor{deepred}{13.6}}/\textbf{\textcolor{deepred}{24.07}}/\textbf{\textcolor{blue}{33.5}}& \textbf{\textcolor{deepred}{13.5}}/\textbf{\textcolor{deepred}{27.7}}/\textbf{\textcolor{deepred}{35.51}}&                                         \textbf{\textcolor{deepred}{15.74}}/\textbf{\textcolor{deepred}{30.02}}/\textbf{\textcolor{deepred}{39.65}}\\ \bottomrule
\end{tabular}
\label{tab:task-wise-20-4}
\end{table*}

\begin{table*}[]
\centering
\caption{20-2 results on YouTube-VIS-2019 dataset.}
\begin{tabular}{>{\centering\arraybackslash}p{0.13\linewidth}>{\centering\arraybackslash}p{0.05\linewidth}>{\centering\arraybackslash}p{0.05\linewidth}>{\centering\arraybackslash}p{0.05\linewidth}>{\centering\arraybackslash}p{0.05\linewidth}>{\centering\arraybackslash}p{0.05\linewidth}>{\centering\arraybackslash}p{0.05\linewidth}>{\centering\arraybackslash}p{0.05\linewidth}>{\centering\arraybackslash}p{0.05\linewidth}>{\centering\arraybackslash}p{0.05\linewidth}>{\centering\arraybackslash}p{0.05\linewidth}>{\centering\arraybackslash}p{0.05\linewidth}}\toprule
   \textbf{Model}  & \textbf{Step0}& \textbf{Step1}& \textbf{Step2}& \textbf{Step3}& \textbf{Step4}&\textbf{Step5}& \textbf{Step6}& \textbf{Step7}& \textbf{Step8}& \textbf{Step9}&\textbf{Step10}\\ \midrule
                                                   \makecell[c]{FT}& \makecell[c]{12.2\\17.64\\21.07}& \makecell[c]{\textbf{\textcolor{blue}{12.21}}\\15.42\\17.19}& \makecell[c]{3.88\\11.53\\10.95}& \makecell[c]{ 3.71\\9.51\\10.73}& \makecell[c]{ 5.44\\8.8\\10.18}& \makecell[c]{    3.38\\8.66\\11.29}& \makecell[c]{ 6.71\\3.95\\10.26}& \makecell[c]{ 5.79\\2.35\\6.71}& \makecell[c]{ 2.29\\3.52\\4.56}& \makecell[c]{ 6.34\\3.09\\6.24}& \makecell[c]{1.11\\2.85\\7.22}\\ \midrule
                                                   \makecell[c]{MiB\cite{Cer2020MiB}}             & \makecell[c]{10.82\\16.63\\22.15}& \makecell[c]{ 11.78\\14.13\\17.57}& \makecell[c]{ 10.09\\12.97\\18.81}& \makecell[c]{ 7.13\\10.93\\18.77}& \makecell[c]{ 7.36\\10.21\\16.23}& \makecell[c]{ 7.68\\9.16\\16.28}& \makecell[c]{ 4.62\\8.79\\14.97}& \makecell[c]{ 1.12\\2.61\\9.26}& \makecell[c]{ 1.32\\2.18\\6.09}& \makecell[c]{ 3.51\\2.09\\5.43}& \makecell[c]{0.37\\3.1\\3.94}\\ \midrule
                                                   \makecell[c]{CoMFormer\cite{cermelli2023comformer}}       & \makecell[c]{8.53\\13.78\\17.81}& \makecell[c]{ 7.73\\11.95\\17.38}& \makecell[c]{ 9.00\\12.63\\19.85}& \makecell[c]{ 6.93\\11.65\\19.13}& \makecell[c]{6.29\\7.36\\17.35}&\makecell[c]{3.95\\.7.61\\15.30}& \makecell[c]{5.69\\5.01\\15.90}& \makecell[c]{4.43\\4.28\\10.94}& \makecell[c]{1.96\\3.81\\11.52}& \makecell[c]{2.79\\6.25\\11.26\\}&\makecell[c]{2.51\\4.77\\9.52}\\ \midrule
                                                   \makecell[c]{ECLIPSE\cite{kim2024eclipse}}         & \makecell[c]{\textbf{\textcolor{blue}{12.42}}\\\textbf{\textcolor{blue}{22.52}}\\27.58}& \makecell[c]{11.45\\\textbf{\textcolor{blue}{21.1}}\\25.89}& \makecell[c]{\textbf{\textcolor{blue}{10.92}}\\\textbf{\textcolor{deepred}{21.8}}\\\textbf{\textcolor{blue}{26.77}}}& \makecell[c]{\textbf{\textcolor{blue}{9.6}}\\\textbf{\textcolor{blue}{20.44}}\\\textbf{\textcolor{blue}{25.84}}}& \makecell[c]{\textbf{\textcolor{blue}{9.51}}\\\textbf{\textcolor{blue}{19.55}}\\\textbf{\textcolor{blue}{26.49}}}& \makecell[c] {\textbf{\textcolor{blue}{9.51}}\\\textbf{\textcolor{blue}{19.4}}\\\textbf{\textcolor{blue}{27.91}}}& \makecell[c]{\textbf{\textcolor{blue}{10.78}}\\\textbf{\textcolor{blue}{17.98}}\\\textbf{\textcolor{blue}{29.26}}}& \makecell[c]{\textbf{\textcolor{blue}{10.03}}\\\textbf{\textcolor{blue}{17.58}}\\\textbf{\textcolor{blue}{28.81}}}& \makecell[c]{\textbf{\textcolor{blue}{10.02}}\\\textbf{\textcolor{blue}{17.57}}\\\textbf{\textcolor{blue}{27.5}}}& \makecell[c]{\textbf{\textcolor{blue}{10.64}}\\\textbf{\textcolor{blue}{16.8}}\\\textbf{\textcolor{blue}{26.82}}}&\makecell[c]{\textbf{\textcolor{blue}{10.58}}\\\textbf{\textcolor{blue}{17.28}}\\\textbf{\textcolor{blue}{28.58}}}\\ \midrule
                                                   \makecell[c]{CoMBO\cite{fang2025combo}}           & \makecell[c]{10.77\\21.10\\\textbf{\textcolor{blue}{28.06}}}&\makecell[c]{11.08\\\textbf{\textcolor{deepred}{22.01}}\\\textbf{\textcolor{deepred}{29.72}}}&\makecell[c]{7.88\\20.51\\26.08}& \makecell[c]{8.65\\18.51\\24.70}&\makecell[c]{4.79\\15.40\\22.60}& \makecell[c]{7.41\\16.32\\22.58}& \makecell[c]{6.31\\9.00\\17.64}& \makecell[c]{4.50\\5.85\\16.67}& \makecell[c]{3.56\\5.38\\13.52}& \makecell[c]{4.26\\2.92\\10.21}&\makecell[c]{2.91\\3.26\\8.29}\\ \midrule
 \makecell[c]{CRISP} & \makecell[c]{\textbf{\textcolor{deepred}{13.53}}\\ \textbf{\textcolor{deepred}{22.99}} \\\textbf{\textcolor{deepred}{29.33}}}& \makecell[c]{\textbf{\textcolor{deepred}{12.67}}\\20.94\\\textbf{\textcolor{blue}{28.68}}}& \makecell[c]{\textbf{\textcolor{deepred}{12.71}}\\\textbf{\textcolor{blue}{21.57}}\\\textbf{\textcolor{deepred}{29.31}}}& \makecell[c]{\textbf{\textcolor{deepred}{12.71}}\\\textbf{\textcolor{deepred}{22.18}}\\\textbf{\textcolor{deepred}{29.60}}}& \makecell[c]{\textbf{\textcolor{deepred}{13.09}}\\\textbf{\textcolor{deepred}{22.18}}\\\textbf{\textcolor{deepred}{29.91}}}& \makecell[c]{\textbf{\textcolor{deepred}{13.09}}\\\textbf{\textcolor{deepred}{22.31}}\\\textbf{\textcolor{deepred}{31.95}}}& \makecell[c]{\textbf{\textcolor{deepred}{15.52}}\\\textbf{\textcolor{deepred}{22.06}}\\\textbf{\textcolor{deepred}{34.34}}}& \makecell[c]{\textbf{\textcolor{deepred}{15.51}}\\\textbf{\textcolor{deepred}{22.06}}\\\textbf{\textcolor{deepred}{34.40}}}& \makecell[c]{\textbf{\textcolor{deepred}{15.51}}\\\textbf{\textcolor{deepred}{23.86}}\\\textbf{\textcolor{deepred}{34.45}}}& \makecell[c]{\textbf{\textcolor{deepred}{15.51}}\\\textbf{\textcolor{deepred}{24.01}}\\\textbf{\textcolor{deepred}{34.51}}}&\makecell[c]{\textbf{\textcolor{deepred}{14.50}}\\\textbf{\textcolor{deepred}{24.59}}\\\textbf{\textcolor{deepred}{34.44}}}\\ \bottomrule
\end{tabular}
\label{tab:task-wise-20-2}
\end{table*}

\begin{table*}[]
\centering
\caption{20-5 results on YouTube-VIS-2021 dataset.}
\begin{tabular}{>{\centering\arraybackslash}p{0.14\linewidth}>{\centering\arraybackslash}p{0.14\linewidth}>{\centering\arraybackslash}p{0.14\linewidth}>{\centering\arraybackslash}p{0.14\linewidth}>{\centering\arraybackslash}p{0.14\linewidth}>{\centering\arraybackslash}p{0.14\linewidth}}\toprule
   \textbf{Model} & \textbf{Step0}& \textbf{Step1}& \textbf{Step2}& \textbf{Step3}& \textbf{Step4}\\ \midrule
                                                   FT             & 8.91/18.53/22.21& 6.21/14.05/20.31& 0.93/2.75/4.66& 0.99/5.12/3.24& 2.42/6.4/5.55\\
                                                   MiB\cite{Cer2020MiB}            & 8.95/17.64/22.74& 4.78/15.57/18.91& 5.76/12.97/18.98& 1.79/13.52/14.65& 2.72/16.68/12.51\\
                                                   CoMFormer \cite{cermelli2023comformer}      & 6.51/12.67/20.96& 6.93/15.25/25.93& 6.41/16.9/25.74& 7.08/18.69/23.82& 8.34/16.68/26.03\\
                                                   ECLIPSE\cite{kim2024eclipse}        & 10.07/22.76/31.62& \textbf{\textcolor{blue}{8.74}}/20.9/31.41& 9.43/23.23/35.07& \textbf{\textcolor{blue}{9.27}}/\textbf{\textcolor{blue}{25.84}}/\textbf{\textcolor{blue}{33.58}}& 9.84/\textbf{\textcolor{blue}{29.87}}/\textbf{\textcolor{blue}{35.26}}\\
                                                   CoMBO\cite{fang2025combo}          &                         \textbf{\textcolor{blue}{10.16}}/\textbf{\textcolor{blue}{23.06}}/\textbf{\textcolor{deepred}{32.11}}&                         8.07/\textbf{\textcolor{blue}{21.26}}/\textbf{\textcolor{deepred}{32.32}}&                         \textbf{\textcolor{blue}{10.25}}/\textbf{\textcolor{deepred}{27.2}}/\textbf{\textcolor{deepred}{38.29}}&                         8.06/22.33/31.60&                         \textbf{\textcolor{blue}{10.06}}/18.22/28.65\\
                          CRISP           & \textbf{\textcolor{deepred}{11.04}}/\textbf{\textcolor{deepred}{24.78}}/\textbf{\textcolor{blue}{31.95}}& \textbf{\textcolor{deepred}{10.2}}/\textbf{\textcolor{deepred}{22.47}}/\textbf{\textcolor{blue}{32.26}}& \textbf{\textcolor{deepred}{10.8}}/\textbf{\textcolor{blue}{24.71}}/\textbf{\textcolor{blue}{36.18}}& \textbf{\textcolor{deepred}{10.97}}/\textbf{\textcolor{deepred}{27.82}}/\textbf{\textcolor{deepred}{35.26}}& \textbf{\textcolor{deepred}{12.88}}/\textbf{\textcolor{deepred}{31.35}}/\textbf{\textcolor{deepred}{39.42}}\\ \bottomrule
\end{tabular}
\label{tab:task-wise-20-5}
\end{table*}

\begin{table}[]
\caption{10-10 results on YouTube-VIS-2021 dataset.}
\centering
\begin{tabularx}{0.95\linewidth}{ 
    >{\raggedright}p{0.23\linewidth}  
    *{4}{>{\centering\arraybackslash}p{0.12\linewidth}} 
}
\toprule
   \makecell[c]{\textbf{Model}} & \makecell[c]{\textbf{Step0}}& \makecell[c]{\textbf{Step1}}& \makecell[c]{\textbf{Step2}}& \makecell[c]{\textbf{Step3}}\\ \midrule
                                                   \makecell[c]{\mbox{FT}}& \makecell[c]{5.96\\10.28\\13.67}& \makecell[c]{6.10\\10.32\\16.46}& \makecell[c]{3.6\\6.31\\11.26}& \makecell[c]{3.39\\8.32\\6.67}\\\midrule
                                                   \makecell[c]{\mbox{MiB\cite{Cer2020MiB}}} & \makecell[c]{6.1\\17.92\\12.35}& \makecell[c]{6.53\\14.46\\19.61}& \makecell[c]{6.31\\16.08\\23.76}& \makecell[c]{5.75\\22.12\\23.68}\\\midrule
                                                   \makecell[c]{CoMFormer\cite{cermelli2023comformer}}& \makecell[c]{5.64\\10.67\\11.90}& \makecell[c]{\textbf{\textcolor{deepred}{8.44}}\\15.65\\21.96}& \makecell[c]{6.48\\18.32\\25.55}& \makecell[c]{6.41\\19.22\\24.56}\\\midrule
                                                   \makecell[c]{ECLIPSE\cite{kim2024eclipse}}        & \makecell[c]{\textbf{\textcolor{blue}{6.59}}\\\textbf{\textcolor{deepred}{12.18}}\\\textbf{\textcolor{deepred}{16.12}}}& \makecell[c]{6.04\\16.54\\23.00}& \makecell[c]{6.03\\20.09\\\textbf{\textcolor{blue}{30.86}}}& \makecell[c]{7.10\\25.37\\\textbf{\textcolor{deepred}{35.47}}}\\\midrule
                                                   \makecell[c]{CoMBO\cite{fang2025combo}}         &                         \makecell[c]{5.62\\11.32\\15.26}&                         \makecell[c]{6.24\\\textbf{\textcolor{blue}{17.91}}\\\textbf{\textcolor{deepred}{24.72}}}&                         \makecell[c]{\textbf{\textcolor{blue}{7.43}}\\\textbf{\textcolor{deepred}{23.5}}\\\textbf{\textcolor{deepred}{36.46}}}&                         \makecell[c]{\textbf{\textcolor{blue}{8.39}}\\\textbf{\textcolor{blue}{27.55}}\\32.76}\\\midrule
                          \makecell[c]{\mbox{CRISP}}& \makecell[c]{\textbf{\textcolor{deepred}{6.67}}\\\textbf{\textcolor{blue}{12.02}}\\\textbf{\textcolor{blue}{15.78}}}& \makecell[c]{\textbf{\textcolor{blue}{8.3}}\\\textbf{\textcolor{deepred}{18.86}}\\\textbf{\textcolor{blue}{24.57}}}& \makecell[c]{\textbf{\textcolor{deepred}{8.99}}\\\textbf{\textcolor{blue}{22.6}}\\30.66}& \makecell[c]{\textbf{\textcolor{deepred}{9.6}}\\\textbf{\textcolor{deepred}{28.43}}\\\textbf{\textcolor{blue}{34.2}}}\\ \bottomrule
\end{tabularx}
\label{tab:task-wise-10-10}
\end{table}

\begin{figure*}[]
\centering
\subfloat{\includegraphics[width=7in]{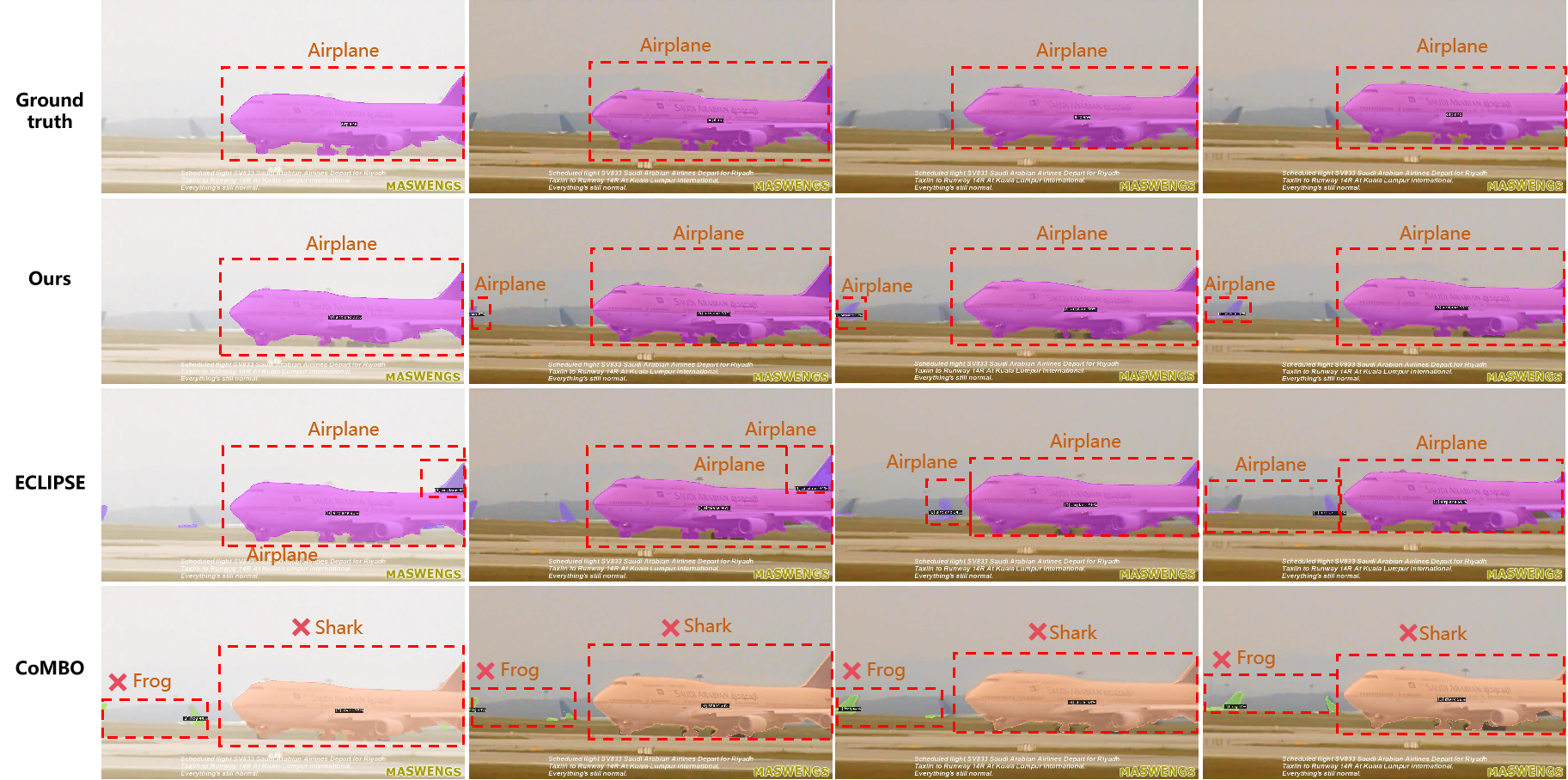}%
\label{fig: 4comparation1}}
\hfil
\subfloat{\includegraphics[width=7in]{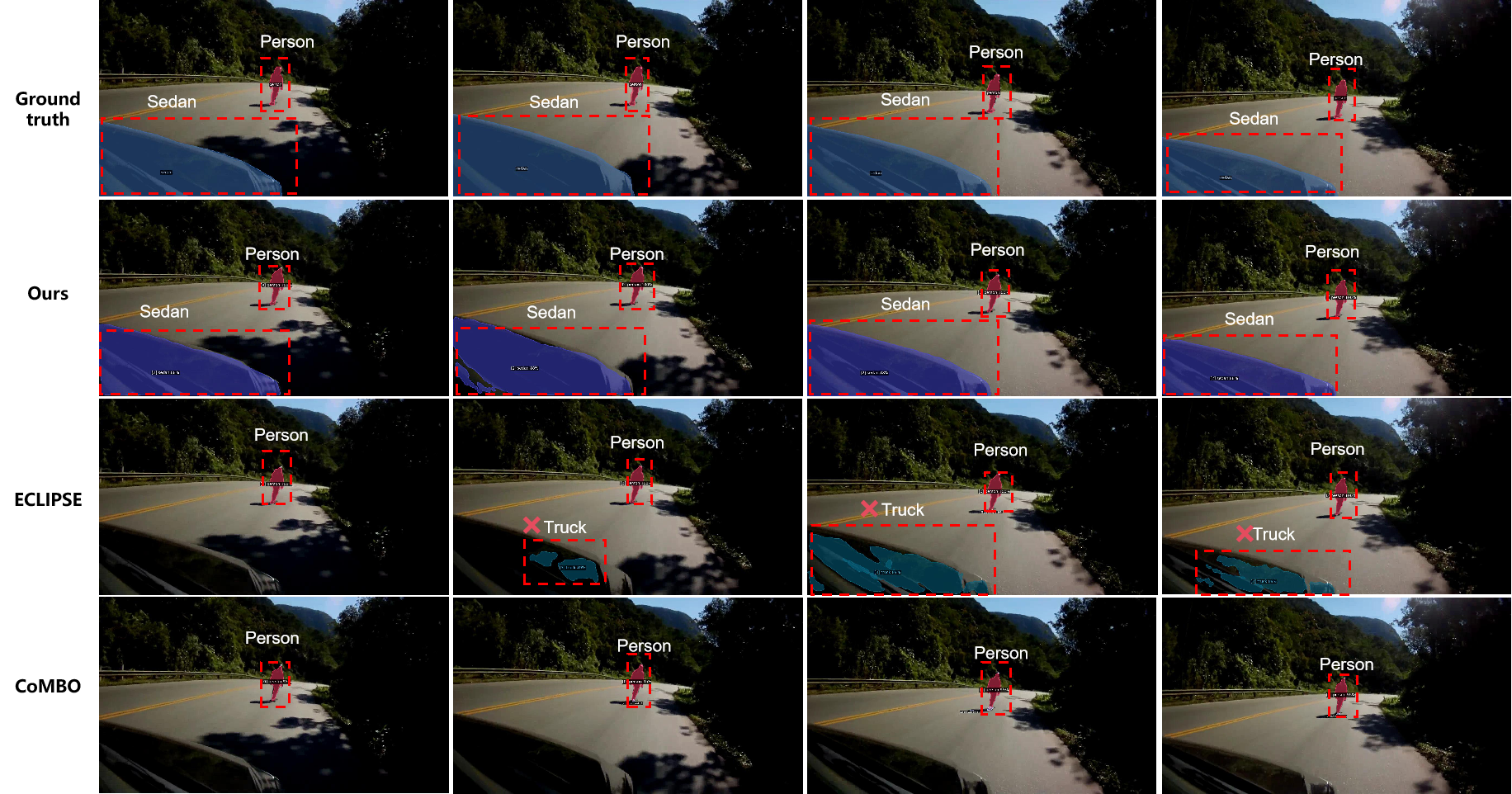}%
\label{fig: 4comparation2}}
\caption{Comparison analysis with other methods. In the upper example, our method accurately segments both the \textbf{person} and the \textbf{sedan} while maintaining consistent instance tracking over time. In the lower example, it correctly identifies the \textbf{airplane} and additionally segments an instance even not annotated in the ground truth.}
\label{fig: qualitative-analysis}
\end{figure*}

\begin{figure*}[!t]
\centering
\includegraphics[width=7in]{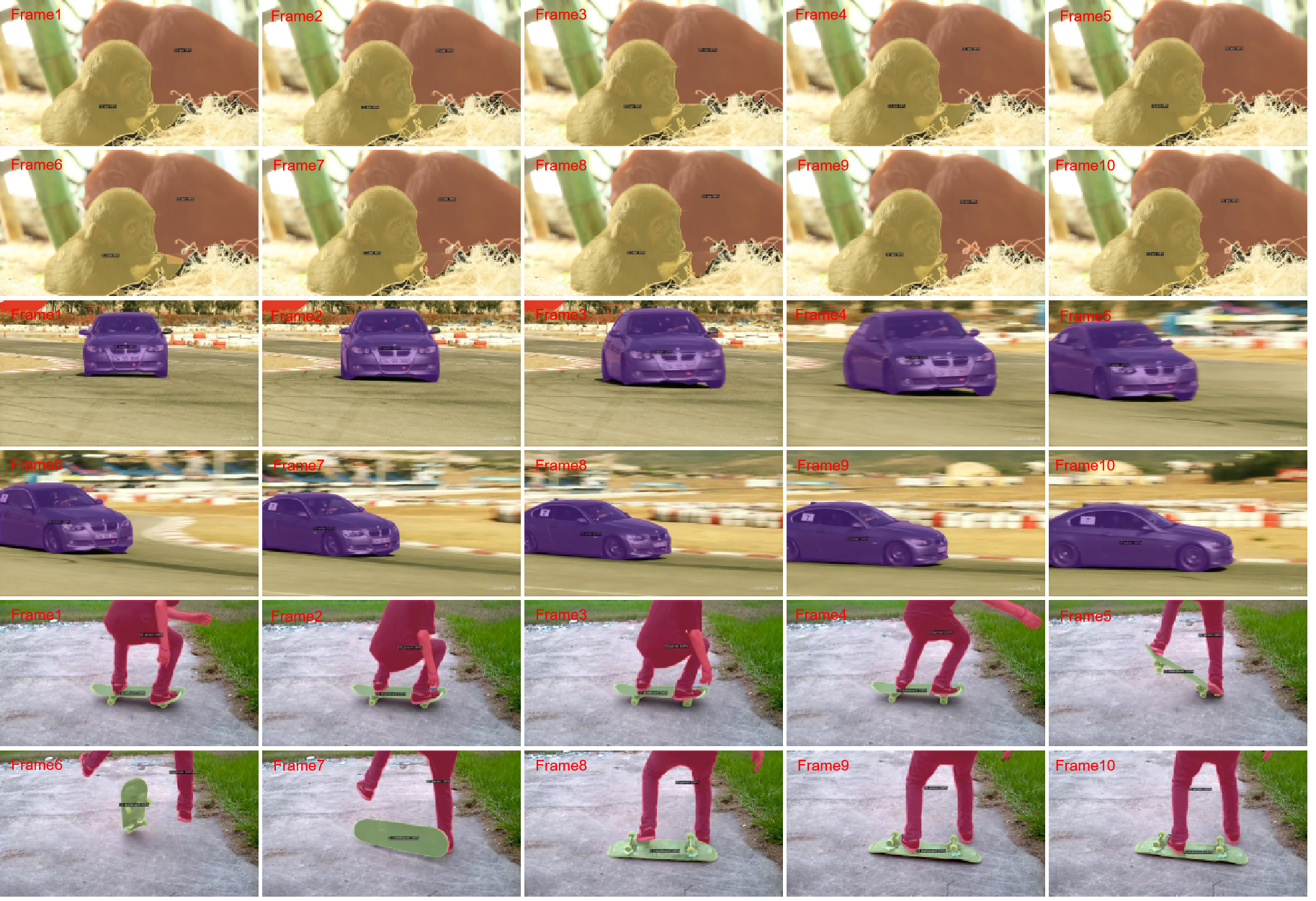}
\caption{Visualization results of several challenging samples by our CRISP. In the first set of video frames, our CRISP successfully achieved precise segmentation of two overlapping apes with similar appearances. The second set of frames demonstrates effective tracking capability of our CRISP for the high-speed moving car. The final experimental results prove that CRISP can maintain stable segmentation performance even when target persons temporarily disappear and reappear.}
\label{fig: other-show}
\end{figure*}

\section{Experiment}

\subsection{Experimental Setting}

\textbf{Datasets:}
We evaluate our approach on YouTube-VIS-2019 \cite{Yang2019vis} and YouTube-VIS-2021 \cite{vis2021}. YouTube-VIS-2019 contains 2,883 high-resolution YouTube videos, including 2,238 training videos, 302 validation videos, and 343 test videos. The annotations include 40 common object categories such as humans, animals, and vehicles. YouTube-VIS-2021 has a total of 3,859 high-resolution YouTube videos, with 2,985 training videos, 421 validation videos and 453 test videos. The labels are based on those from YouTube-VIS-2019 but have been updated by merging eagles and owls into birds, combining apes into monkeys, removing hands, and adding new categories, including flying disc, squirrel, and whale.

\textbf{Metrics:}
We report several standard metrics for comprehensive evaluation\cite{Yang2019vis, vis2021} across the two datasets. 
Average Precision (AP) (area under the precision-recall curve), with final-task results reporting mAP (mean average precision), AP50 and AP75, while per-task analysis includes APs, APm and APl.
Average Recall (AR) (maximum recall under fixed segmented instances per video), reporting AR1 and AR10; and for continual learning evaluation.
Forgetting Ratio (FR) quantifying how much knowledge from previous tasks is forgotten during new task learning. The calculation of FR follows:
\begin{equation}
    \label{forgetting}
   FR=\frac{1}{N_c}\!\sum^{T-1}_{t=1}\frac{1}{T\!-\!t} \!\sum^{C_t}_{c_t=1} \frac{\mathbb{I}(A_{t,c}\!-\!A_{T,c})_{\ge 0}(A_{t,c}\!-\!A_{T,c})}{A_{t,c}},
\end{equation}
where $N_c$ represents the number of categories, $C_t$ represents the number of categories in task $t$, $A_{t,c}$ and $A_{T,c}$ represent the mAP of the model when it first learns category $c$ and the final task, respectively. $\mathbb{I}(A_{t,c}-A_{T,c})_{\ge 0}$ can be denoted as:
\begin{equation}
    \label{I4forget}
      \mathbb{I}_{ij} =
        \begin{cases}
        0 & \text{if } A_{t,c}-A_{T,c} \ge 0 \\
        1 & \text{otherwise.}
        \end{cases}
\end{equation}

\textbf{Continual Learning Protocol:}
We adopt the incremental methodology from prior tasks and characterize the scenarios using the notation $N_{ini}-N_{inc}$, where $N_{ini}$ signifies the initial class count and $N_{inc}$ represents the number of classes added during each incremental phase. 
For comprehensive evaluation, we design two challenging protocols for each dataset.
For YouTube-VIS-2019, we set \textbf{20-4}, which learns 20 base classes first, then 4 new classes per step ($N_t$=6) and  \textbf{20-2}, which learns 20 base classes, then 2 new classes per step ($N_t$=11).
For YouTube-VIS-2021, we set \textbf{20-5}, which learns 20 base classes first, then 5 new classes per step ($N_t$=5) and \textbf{10-10}, which learns 10 base classes, then 10 new classes per step ($N_t$=4).

\begin{figure*}[!t]
\centering
\subfloat[]{\includegraphics[width=1.7in]{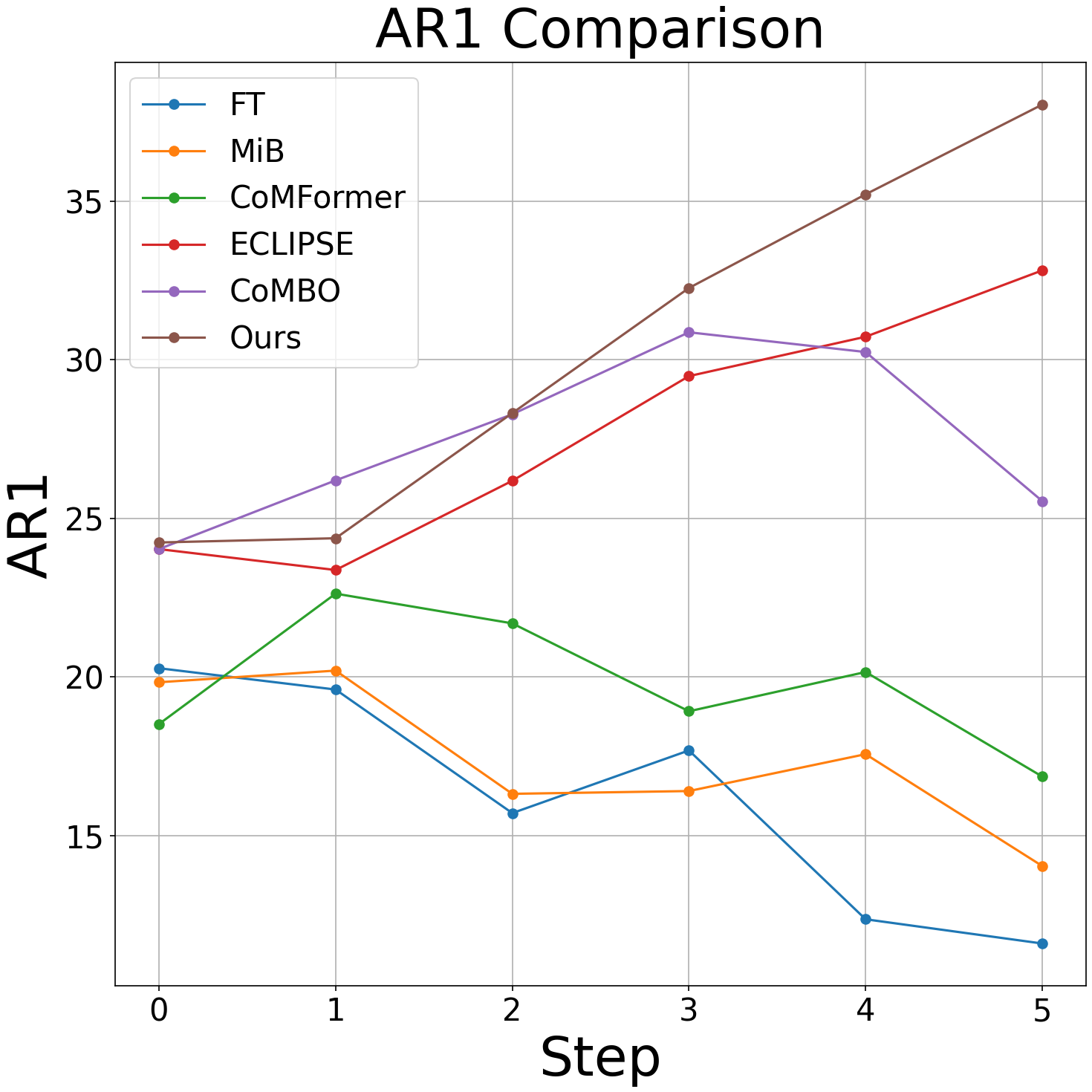}%
\label{fig_ar1_20-4}}
\subfloat[]{\includegraphics[width=1.7in]{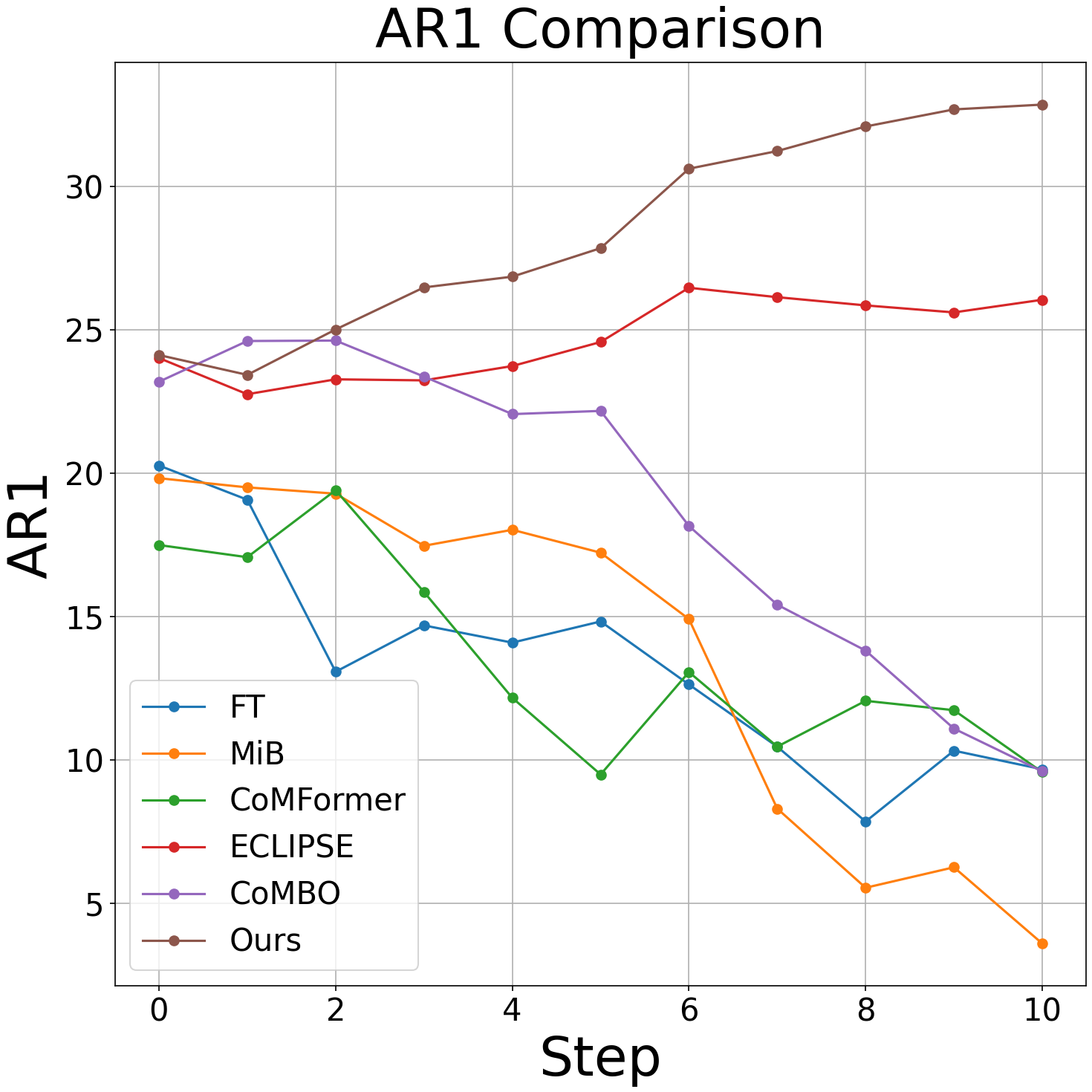}%
\label{fig_ar1_20-2}}
\subfloat[]{\includegraphics[width=1.7in]{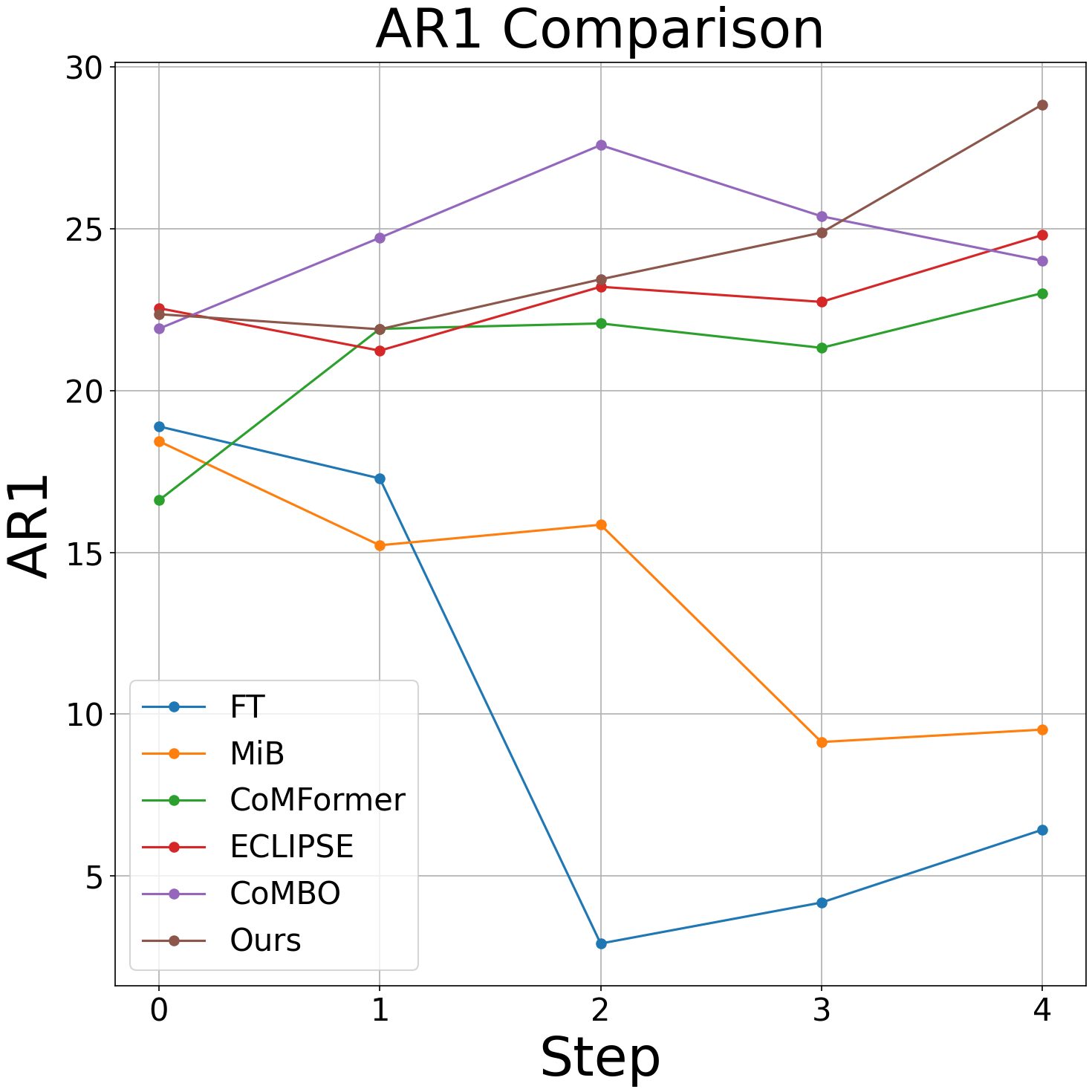}%
\label{fig_ar1_20-5}}
\subfloat[]{\includegraphics[width=1.7in]{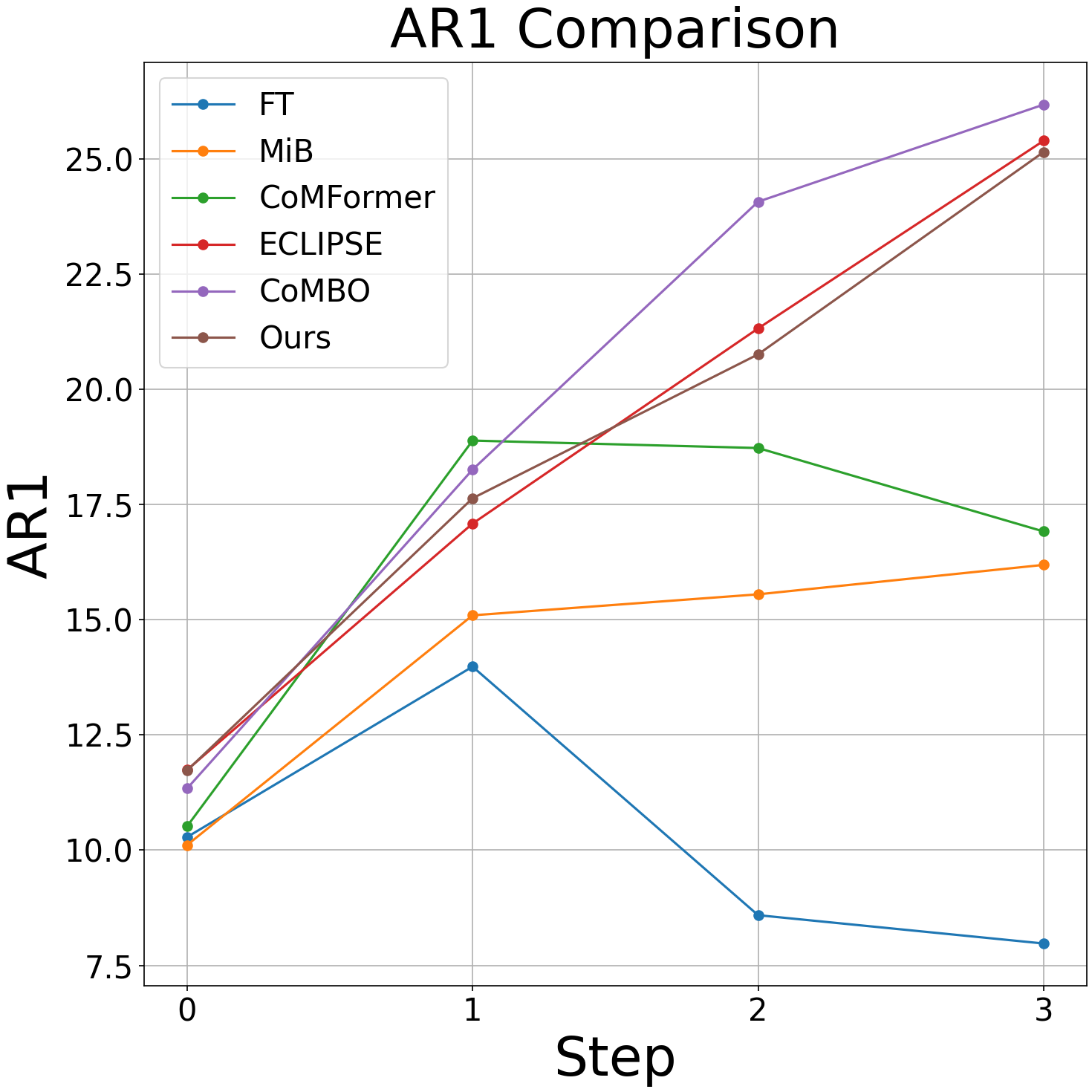}%
\label{fig_ar1_10-10}}
\caption{AR1 results of each step of 20-4, 20-2 scenarios on YouTube-VIS-2019 dataset and 20-5, 10-10 scenarios on Youtube-VIS-2021.}
\label{fig: AR1-task-wise}
\end{figure*}

\begin{figure*}[!t]
\centering
\subfloat[]{\includegraphics[width=1.7in]{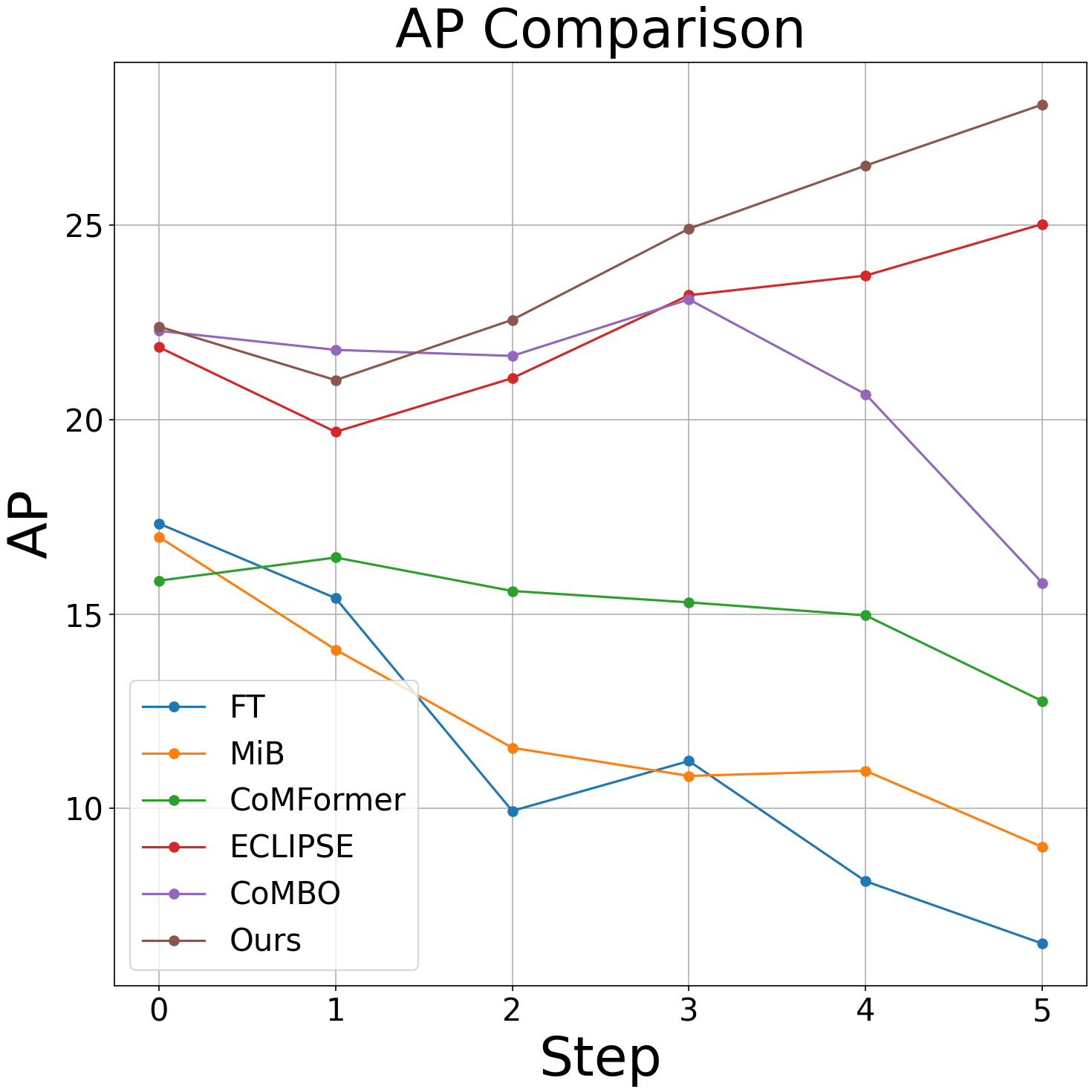}%
\label{fig_ap_20-4}}
\subfloat[]{\includegraphics[width=1.7in]{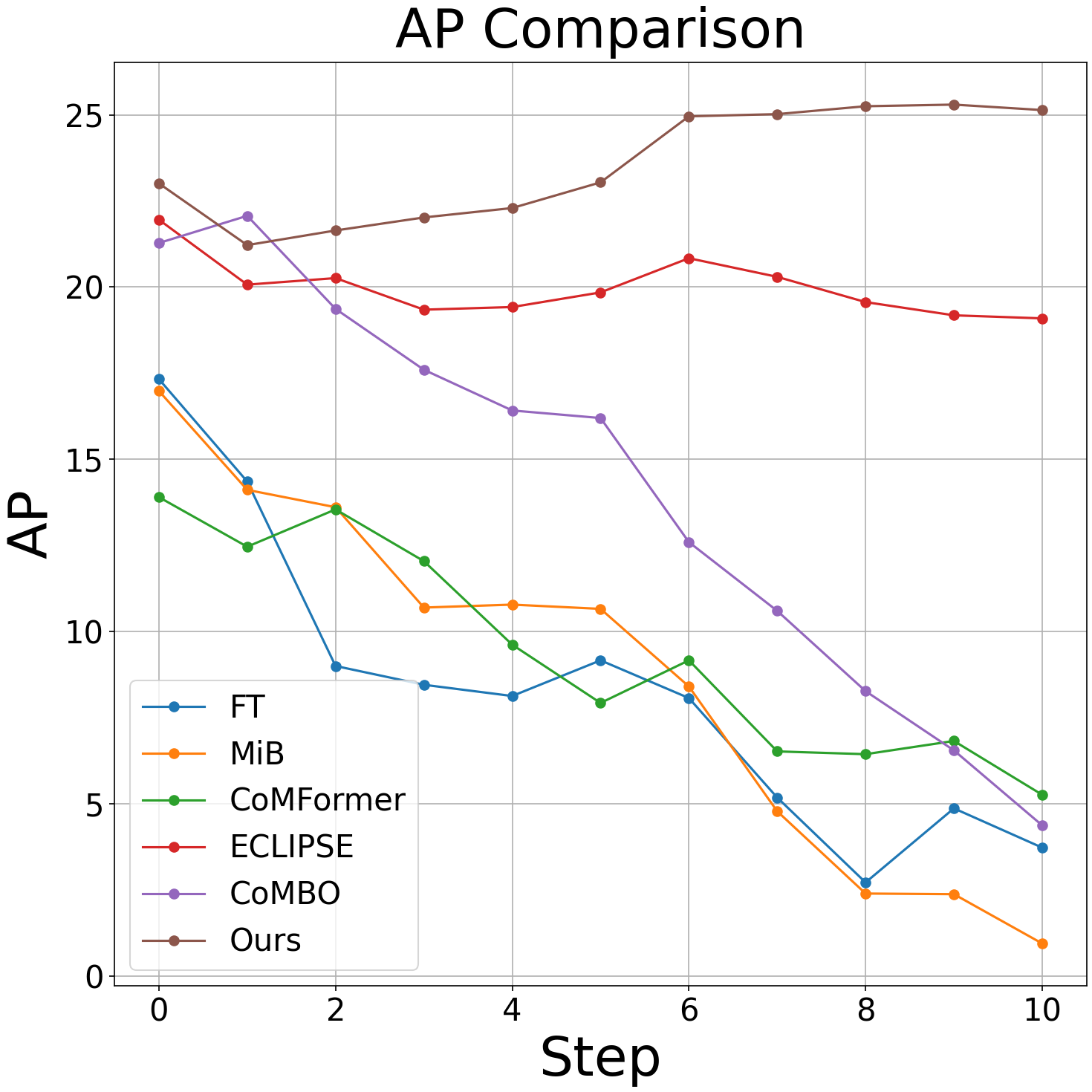}%
\label{fig_ap_20-2}}
\subfloat[]{\includegraphics[width=1.7in]{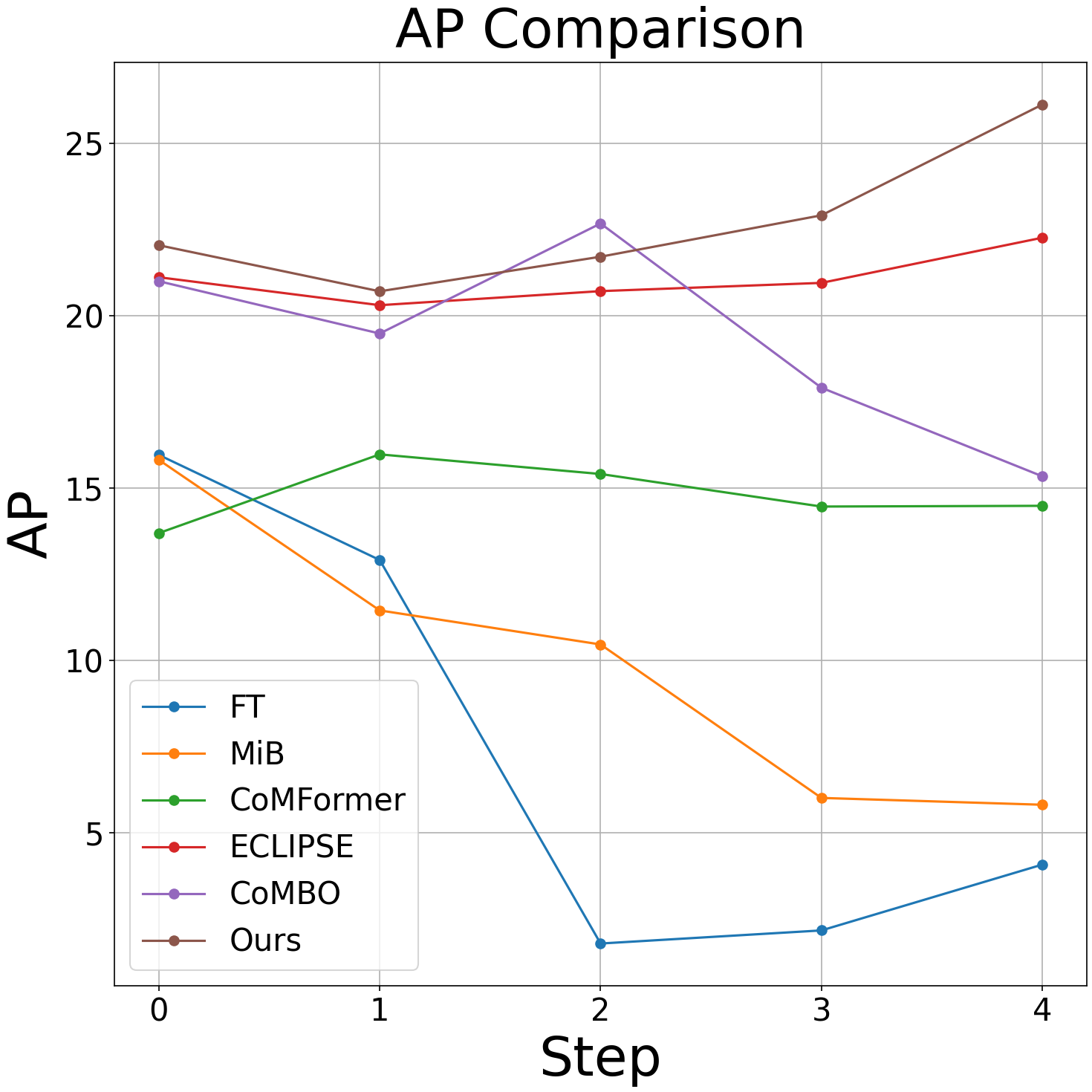}%
\label{fig_ap_20-5}}
\subfloat[]{\includegraphics[width=1.7in]{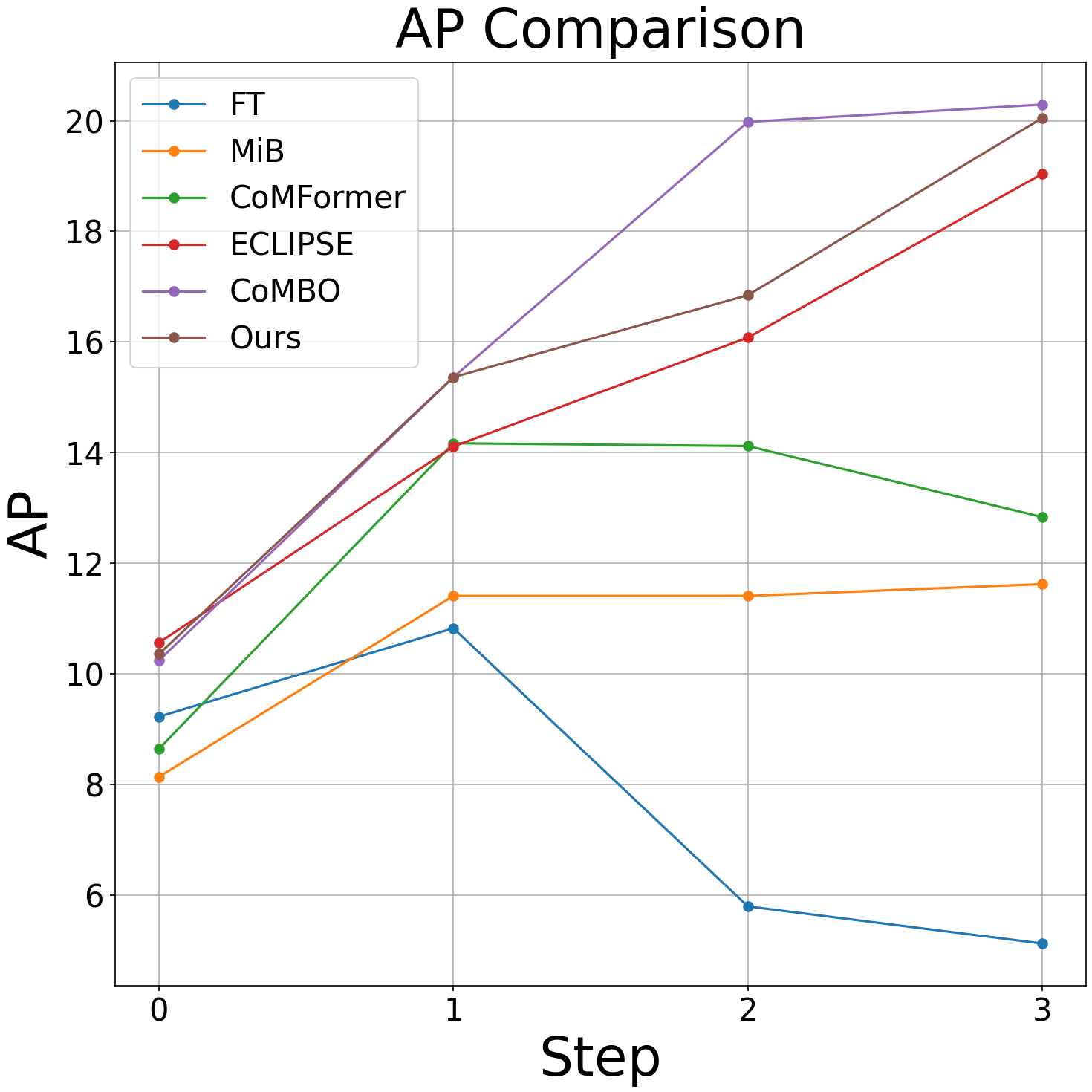}%
\label{fig_ap_10-10}}
\caption{mAP results of each step of 20-4, 20-2 scenarios on YouTube-VIS-2019 dataset and 20-5, 10-10 scenarios on Youtube-VIS-2021.}
\label{fig: AP-task-wise}
\end{figure*}

\textbf{Training Settings:}
We develop our CRISP approach for CVIS, building upon the Mask2Former framework with ResNet-50 as the backbone. We adapt training parameters according to dataset and task characteristics. For YouTube-VIS-2019, the initial task is trained with a learning rate of $0.0001$ and a batch size of $16$ for $1500$ iterations. During incremental learning, each class is fine-tuned for $150$ iterations. For the 20-5 task on YouTube-VIS-2021, the same base parameters are used in the initial phase, while the incremental iterations per class are increased to $300$. For the 10-10 task, the number of initial iterations is reduced to $750$, with the other parameters unchanged.

\subsection{Experimental Results}
We compare our approach with MiB\cite{Cer2020MiB}, CoMFormer \cite{cermelli2023comformer}, ECLIPSE\cite{kim2024eclipse}, CoMBO\cite{fang2025combo}, and Fine-tune (FT). FT stands for training the model without any continual learning methods, which demonstrates the fundamental performance. As summarized in Tab. \ref{tab:exp_table0} and Tab. \ref{tab:exp_table}, we achieved the state-of-the-art perference on three scenarios and second place in one scenarios.

\textbf{Qualitative Comparison:}
Fig. \ref{fig: qualitative-analysis} demonstrates the qualitative analysis results of our method compared to other methods. In the upper example, our method accurately segments both the person and the sedan while preserving consistent instance IDs across the entire video sequence. In the lower example, it successfully detects the airplane and even identifies an additional instance that is absent from the ground-truth annotations, demonstrating its robustness to incomplete labeling. Fig. \ref{fig: other-show} presents visualization results of several challenging cases processed by our CRISP framework, including overlapping similar instances, dynamic scenes, and scenarios where instance pixels disappear and later reappear. In highly similar scenarios with instance overlapping, feature similarity often leads to instance-wise confusion in segmentation. However, CRISP consistently preserves accurate segmentation and classification for these instances. In dynamic scenes, continuous changes in instance shape and appearance may hinder reliable tracking, yet our method maintains stable tracking of cars across frames. Even in cases where an instance partially disappears and later reappears, CRISP demonstrates strong robustness. Notably, it not only predicts accurate masks and categories for instances in videos, even when certain instances are absent from ground-truth annotations, but also delivers superior tracking performance throughout the video sequence.


\textbf{Quantitative Results:}
Tab. \ref{tab:task-wise-20-4}-\ref{tab:task-wise-10-10} present the task-wise APs, APm, and APl, which evaluate the ability to segment small, medium and large objects, respectively. By jointly modeling at both instance-wise and task-wise levels, our approach consistently outperforms distillation-based methods throughout the incremental learning process. In the short-term incremental scenario (10-10), we observed that our method achieved superior performance over CoMBO \cite{fang2025combo} in metrics such as APs, but slightly lagged behind overall mAP on the final task. We hypothesize that this discrepancy stems from the misalignment between the annotated mask boundaries and the true object contours.

Fig. \ref{fig: AR1-task-wise} and Fig. \ref{fig: AP-task-wise} present AR1 and mAP results in all incremental steps. In the long-term incremental scenario, our method achieves state-of-the-art performance. In contrast, under the short-term incremental setting, CoMBO\cite{fang2025combo} shows stronger results. We attribute this to the increased data volume and shorter task intervals in the short-term scenario, which facilitate continual adaptation and reduce the risk of forgetting previously learned tasks.

\begin{table}[t!]
\centering
\caption{Ablation Study on Proposed Components. ARSP: adaptive residual semantic prompt, ISC: instance semantic consistency, PI: PCA-guided initialization, IC: instance correlation loss.}
\begin{tabular}{>{\centering\arraybackslash}p{0.05\linewidth}>{\centering\arraybackslash}p{0.05\linewidth}>{\centering\arraybackslash}p{0.05\linewidth}>{\centering\arraybackslash}p{0.05\linewidth}>{\centering\arraybackslash}p{0.05\linewidth}>{\centering\arraybackslash}p{0.05\linewidth}>{\centering\arraybackslash}p{0.05\linewidth}>{\centering\arraybackslash}p{0.05\linewidth}>{\centering\arraybackslash}p{0.05\linewidth}>{\centering\arraybackslash}p{0.05\linewidth}}
\midrule
 \multirow{2}{*}{ARSP} & \multirow{2}{*}{ISC}& \multirow{2}{*}{PI}& \multirow{2}{*}{IC} & \multicolumn{6}{c}{20-4 on YouTube-VIS-2019
(6 steps)} \\
                       &                           &                      &                     & mAP   & AP50   & AP75   & AR1  & AR10  & FR  \\ \midrule
 &                           &                      &                     &       25.03&        39.83&        27.59&      32.81&       40.48&     2.23\\
 & & $\checkmark$& & 26.3& 40.81& 29.15& 34.71& 41.99&
2.11\\
 & & & $\checkmark$& 25.69& 40.43& 28.35& 33.40& 39.47&2.05\\
                       $\checkmark$& &                      &                     &       26.42&        40.86&        29.48&      33.37&       39.64&     2.06\\
 $\checkmark$                   & & $\checkmark$                  & $\checkmark$                 & 27.64& 42.69& 31.24& 34.84& 42.18&2.05\\
 $\checkmark$                   &                           $\checkmark$& &                     &       27.83&        42.7&        30.97&      34.92&       41.62&     2.04\\
 $\checkmark$                   & $\checkmark$                       & $\checkmark$                  &                     &       27.5&        42.41&        30.30&      36.68&       43.35&     2.27\\
 $\checkmark$                   & $\checkmark$                       & $\checkmark$                  & $\checkmark$                 &       \textbf{\textcolor{deepred}{28.1}}&        \textbf{\textcolor{deepred}{43.3}}&        \textbf{\textcolor{deepred}{31.98}}&      \textbf{\textcolor{deepred}{38.04}}&       \textbf{\textcolor{deepred}{44.71}}&     \textbf{\textcolor{deepred}{1.93}}\\ \bottomrule
\end{tabular}

\label{tab:ablation_study}
\end{table}

\begin{table}[t!]
\centering
\caption{Ablation Study of task-wise on Proposed Components. ARSP: adaptive residual semantic prompt, ISC: instance semantic consistency, PI: PCA-guided initialization, IC: instance correlation loss.}
\begin{tabular}{>{\centering\arraybackslash}p{0.05\linewidth}>{\centering\arraybackslash}p{0.05\linewidth}>{\centering\arraybackslash}p{0.05\linewidth}>{\centering\arraybackslash}p{0.05\linewidth}>{\centering\arraybackslash}p{0.05\linewidth}>{\centering\arraybackslash}p{0.05\linewidth}>{\centering\arraybackslash}p{0.05\linewidth}>{\centering\arraybackslash}p{0.05\linewidth}>{\centering\arraybackslash}p{0.05\linewidth}>{\centering\arraybackslash}p{0.05\linewidth}}
\midrule
 \multirow{2}{*}{ARSP} & \multirow{2}{*}{ISC}& \multirow{2}{*}{PI}& \multirow{2}{*}{IC} & \multicolumn{6}{c}{mAP results of task-wise
(6 steps)} \\
                       &                           &                      &                     & Step0& Step1& Step2& Step3& Step4& Step5\\ \midrule
 &                           &                      &                     &       21.86&        19.69&        21.07&      23.2&       23.71&     25.03\\
 & & $\checkmark$& & 21.86& 20.08& 21.68& 23.51& 24.65&
26.3\\
 & & & $\checkmark$& 21.86& 19.74& 21.1& 23.45& 23.96&25.69\\
                       $\checkmark$& &                      &                     &       22.21&        20.4&        21.78&      24.53&       25.11&     26.42\\
 $\checkmark$                   & & $\checkmark$                  & $\checkmark$                 & 22.21& 20.37& 21.81& 24.53& 25.87&27.65\\
 $\checkmark$                   &                           $\checkmark$& &                     &       \textbf{\textcolor{deepred}{22.4}}&        20.82&        22.47&      25.71&       \textbf{\textcolor{deepred}{26.55}}&     27.83\\
 $\checkmark$                   & $\checkmark$                       & $\checkmark$                  &                     &       \textbf{\textcolor{deepred}{22.4}}&        \textbf{\textcolor{deepred}{21.04}}&        22.47&      24.5&       25.88&     27.5\\
 $\checkmark$                   & $\checkmark$                       & $\checkmark$                  & $\checkmark$                 &       \textbf{\textcolor{deepred}{22.4}}&        \textbf{\textcolor{blue}{21.02}}&        \textbf{\textcolor{deepred}{22.57}}&      \textbf{\textcolor{deepred}{24.91}}&       \textbf{\textcolor{blue}{26.53}}&     \textbf{\textcolor{deepred}{28.1}}\\ \bottomrule
\end{tabular}

\label{tab:ablation_study_task_wise}
\end{table}

\subsection{Ablation Study}

As shown in Tab. \ref{tab:ablation_study}, we conduct an ablation study under the 20–4 split of YouTube-VIS-2019 dataset to quantify the contribution of each proposed component. Removing PCA-guided initialization (PI) causes a clear performance drop, verifying its role in mitigating task-wise confusion by providing better-aligned query initialization. Instance correlation (IC) loss alone improves tracking-related metrics, but its benefits are markedly amplified when combined with PI (Cols. 4–5 vs. Cols. 6–9), highlighting the synergy between inter-task correlation preservation and instance-level discrimination. Adaptive Residual Semantic Prompt (ARSP) further enhances segmentation accuracy through adaptive query–prompt alignment, while instance semantic consistency (ISC) loss strengthens intra-class compactness and inter-class separability, delivering consistent gains across metrics. Notably, the full CRISP model, integrating ARSP, ISC, PI, and IC, achieves the highest overall mAP (28.1) and best per-class performance, demonstrating the complementary nature of all components.

Crucially, Tab. \ref{tab:ablation_study_task_wise} reinforces these findings from a task-wise perspective. When evaluated by the mAP of each learning step, PI consistently alleviates performance degradation in new tasks, IC loss stabilizes instance tracking across temporal contexts, and the joint application of PI and IC preserves both new-task adaptability and old-task retention. This dual confirmation, across both overall and task-wise metrics, provides strong evidence of the robustness and generality of the proposed modules.

\section{Conclusion}

In this work, we propose CRISP, an earlier attempt for continual video instance segmentation task. For category-wise, we inject semantic information into the model through ARSP to guide the model to learn semantic signals during the training process. At the same time, through an instance semantic consistency loss, the model continuously strengthens alignment between query space and semantic space. For task-wise, we employed PCA-guided initialization to preserve the intrinsic connections among tasks while conducting the initialization process. For instance-wise, we introduced an instance correlation loss, which emphasizes the relevance to the previous query space while enhancing the specificity of the current task query. Extensive experiments on YouTube-VIS-2019 and YouTube-VIS-2021 datasets demonstrate that CRISP achieves state-of-the-art performance on continual video instance segmentation tasks for long terms, effectively avoids catastrophic forgetting, and demonstrates its continual learning ability. Currently, our method has only been tested on continual video instance segmentation tasks. Future work will explore more continual video segmentation scenarios, such as continual video panoptic segmentation.

{
    \small
    \bibliographystyle{IEEEtran.bst}
    \bibliography{main}
}

\end{document}